\pdfoutput=1
\documentclass[10pt,twocolumn,letterpaper]{article}
\usepackage{iccv}
\usepackage{times}
\usepackage{epsfig}
\usepackage{graphicx}
\usepackage{amsmath}
\usepackage{amssymb}
\usepackage{subcaption}
\usepackage{multirow}

\usepackage[pagebackref=true,breaklinks=true,letterpaper=true,colorlinks,bookmarks=false]{hyperref}

\iccvfinalcopy 

\def\iccvPaperID{19} 

\ificcvfinal\fi
\begin{document}

\title{HumanMeshNet: Polygonal Mesh Recovery of Humans}

\author{Abbhinav Venkat \\
{\tt\small abbhinav.venkat@research.iiit.ac.in}
\and Chaitanya Patel\\
{\tt\small chaitanya.patel@students.iiit.ac.in}
\and Yudhik Agrawal \\
{\tt\small yudhik.agrawal@research.iiit.ac.in}
\and Avinash Sharma \\
{\tt\small asharma@iiit.ac.in} \\
International Institute of Information Technology, Hyderabad
}

\maketitle

\begin{abstract}
3D Human Body Reconstruction from a monocular image is an important problem in computer vision with applications in virtual and augmented reality platforms, animation industry, en-commerce domain, etc. While several of the existing works formulate it as a volumetric or parametric learning with complex and indirect reliance on re-projections of the mesh, we would like to focus on implicitly learning the mesh representation. To that end, we propose a novel model, HumanMeshNet, that regresses a template mesh's vertices, as well as receives a regularization by the 3D skeletal locations in a multi-branch, multi-task setup.
The image to mesh vertex regression is further regularized by the neighborhood constraint imposed by mesh topology ensuring smooth surface reconstruction. The proposed paradigm can theoretically learn local surface deformations induced by body shape variations and can therefore learn high-resolution meshes going ahead.
We show comparable performance with SoA (in terms of surface and joint error) with far lesser computational complexity, modeling cost and therefore real-time reconstructions on three publicly available datasets. We also show the generalizability of the proposed paradigm for a similar task of predicting hand mesh models. Given these initial results, we would like to exploit the mesh topology in an explicit manner going ahead.
\end{abstract}

\section{Introduction}
\label{sec:intro}

Recovering a 3D human body shape from a monocular image is an ill posed problem in computer vision with great practical importance for many applications, including virtual and augmented reality platforms, animation industry, e-commerce domain, etc.
%
\begin{figure}[h!]
\begin{subfigure}[b]{0.32\linewidth}
\includegraphics[width=0.8\linewidth]{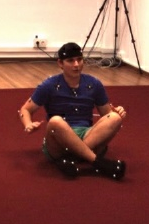}
\end{subfigure}
\begin{subfigure}[b]{0.32\linewidth}
\includegraphics[width=0.8\linewidth]{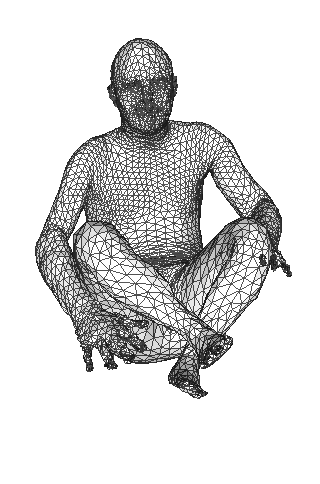}
\end{subfigure}
\begin{subfigure}[b]{0.32\linewidth}
\includegraphics[width=0.8\linewidth]{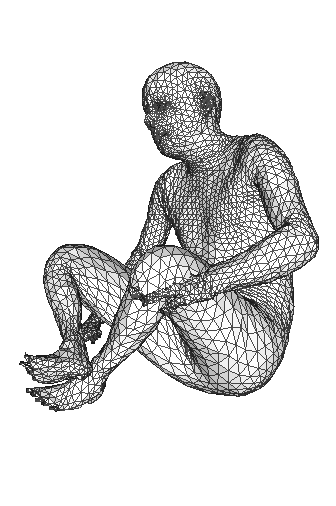}
\end{subfigure}
\begin{subfigure}[b]{0.32\linewidth}
\includegraphics[width=0.8\linewidth]{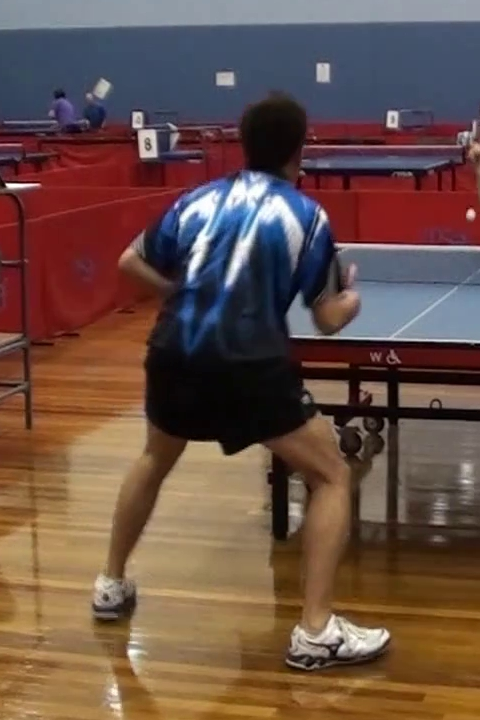}
\end{subfigure}
\begin{subfigure}[b]{0.32\linewidth}
\includegraphics[width=0.8\linewidth]{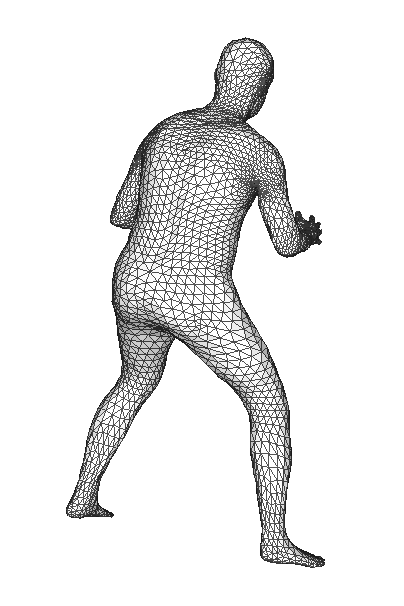}
\end{subfigure}
\begin{subfigure}[b]{0.32\linewidth}
\includegraphics[width=0.8\linewidth]{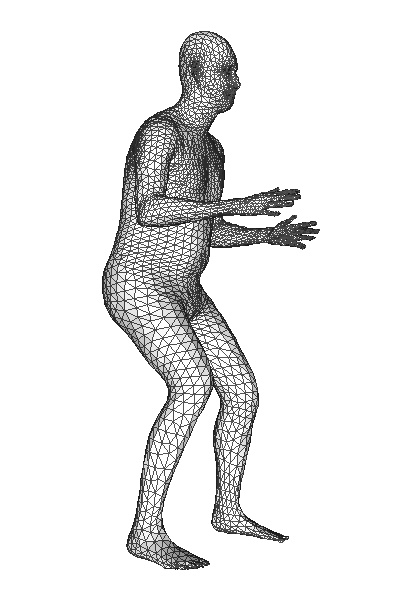}
\end{subfigure}
\caption{We present an early method to integrate Deep Learning with the sparse mesh representation, to successfully reconstruct the 3D mesh of a human from a monocular image}
\label{fig:motivation}
\end{figure}
\begin{figure*}[ht!]
\begin{center}

\includegraphics[width=12.5cm,height=6.0cm]{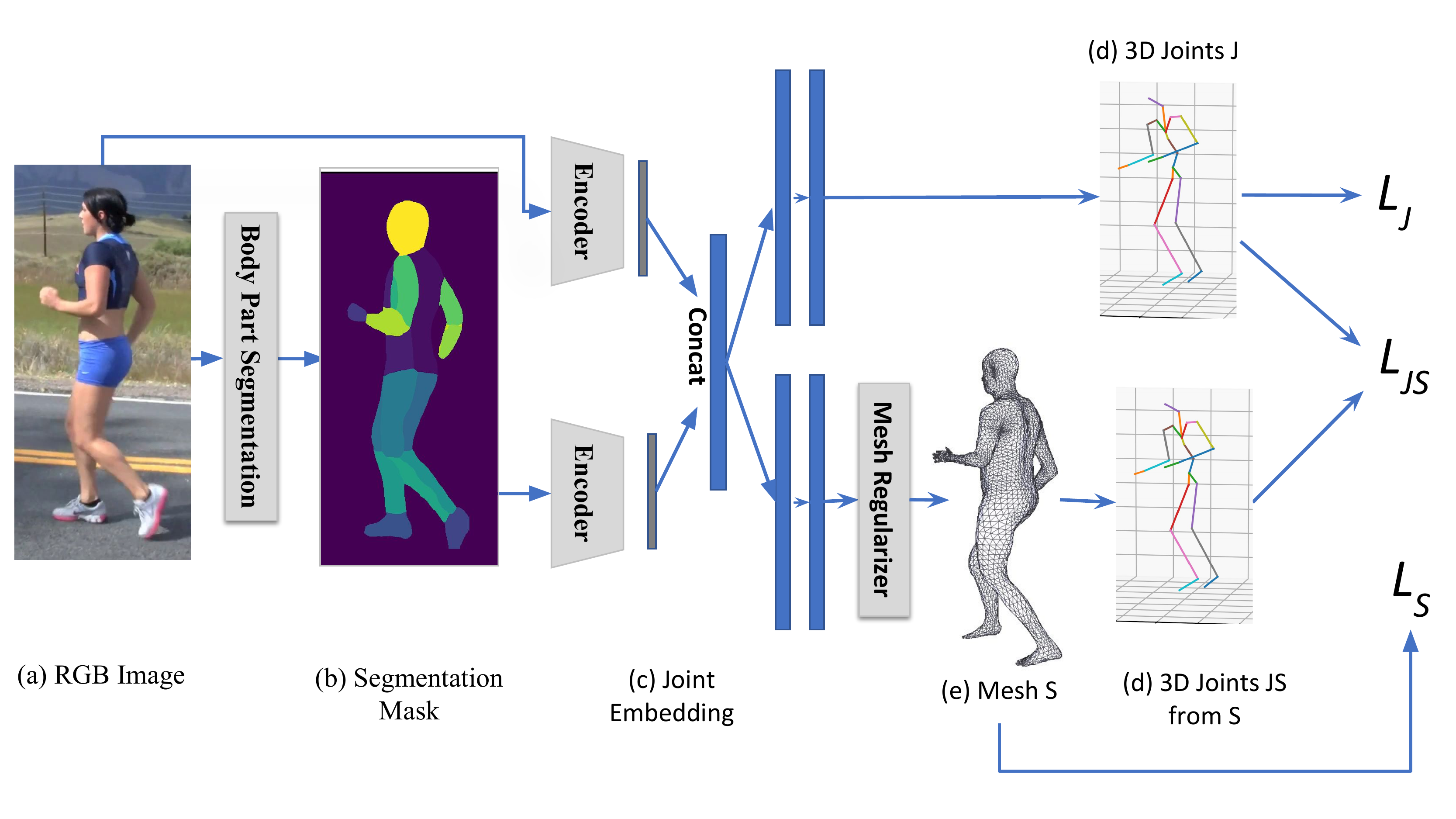}
\caption{Overview of our Multi-Task 3D Human Mesh Reconstruction Pipeline. Given a monocular RGB image  (a),  we first extract a body part-wise segmentation mask using~\cite{alp2018densepose} (b). Then, using a joint embedding of both the RGB and segmentation mask (c), we predict the 3D joint locations (d) and  the 3D mesh (e), in a multi-task setup. The 3D mesh is predicted by first applying a mesh regularizer on the predicted point cloud. Finally, the loss is minimized on both the branches (d) and (e).}
\label{fig:arch}
\end{center}
\end{figure*}
Some of the recent deep learning methods employ volumetric regression to recover the voxel grid reconstruction of human body models from a monocular image~\cite{venkat2018deep,varol2018bodynet}.  Although volumetric regression enables recovering a more accurate surface reconstruction, they do so without an animatable skeleton~\cite{venkat2018deep}, which limits their applicability for some of the aforementioned applications.~\cite{varol2018bodynet} attempted to overcome this limitation by fitting a parametric body model on the volumetric reconstruction using a sillhoute reprojection loss. Nevertheless, in general, such methods yield reconstructions of low resolution at higher computational cost (regression over the cubic voxel grid) and often suffer from broken or partial body parts.

Alternatively, parametric body model~\cite{SCAPETOG2005,SMPL:2015,romero2017embodied} based techniques address some of the above issues, however, at the cost of accurate surface information~\cite{sigal2008combined,guan2009estimating, keepSMPL2016,lassner2017unite}.
Recently, several end-to-end deep learning solutions for estimating the 3D parametric body model from a monocular image have been proposed ~\cite{kanazawa2018end,tan2017indirect,tung2017self,pavlakos2018learning,omran2018neural,xiang2018monocular}. They all attempt to estimate the pose (relative axis-angles) and shape parameters of the SMPL~\cite{SMPL:2015} body model, which is a complex non-linear mapping. To get around this complex mapping, several methods transform them to rotation matrices ~\cite{omran2018neural,pavlakos2018learning} or learn from the 2D/3D keypoint and silhouettes projections (a function of the parameters)~\cite{omran2018neural,kanazawa2018end,pavlakos2018learning}. Additionally, ~\cite{kanazawa2018end} proposes an alternate method for training (Iterative Error Feedback) as well as a body joint specific adversarial losses, which takes upto 5 days to train. In other words, learning the parametric body model hasn't been straightforward and has resulted in complex and indirect solutions that actually rely on different projections of the underlying mesh.

Directly regressing to point cloud or mesh data from image(s) is a severely ill-posed problem and there are very few attempts in deep learning literature in this direction~\cite{RealPoint3D,3DLMNet}. With regard to point cloud regression, most of the attempts have focused on rigid objects, where learning is done in a class specific manner. Apart from a very recent work~\cite{kolotouros2019cmr}, learning a mesh hasn't been explored much for reconstruction primarily because of lack of deep learning constructs to do so.

In this paper, we attempt to work in between a generic point cloud and a mesh - i.e., we learn an "implicitly structured" point cloud. We hypothesize that in order to perform parametric body model based reconstruction, instead of learning the highly non-linear SMPL parameters, learning its corresponding point cloud (although high dimensional) and enforcing the same parametric template topology on it is an easier task. This is because, in SMPL like body models, each of the surface vertices are a sparse linear combination of the transformations induced by the underlying joints i.e., implicitly learning the skinning function by which parametric models are constructed is easier than learning the non-linear axis-angle representation itself (parameters). Further, such models lack high-resolution local surface details as well. Therefore, there are far fewer "representative" points that we have to learn. In comparison with generic point cloud regression as well, this is an easier task because of this implicit structure that exists between these points.

Going ahead, attempting to produce high resolution meshes are a natural extension that is easier in 3D space than in the parametric one. Therefore, we believe that this is a direction worth exploring and we present an initial solution in that direction - HumanMeshNet that simultaneously performs shape estimation by regressing to template mesh vertices (by minimizing surface loss) as well receives a body pose regularisation from a parallel branch in multi-task setup. The image to mesh vertex regression is further explicitly conditioned on the neighborhood constraint imposed by the mesh topology, thus ensuring a smooth surface reconstruction. Figure~\ref{fig:arch} outlines the architecture of HumanMeshNet.

Ours is a relatively simpler model as compared to majority of the existing methods for volumetric and parametric model prediction (e.g.,~\cite{varol2018bodynet}). This makes it efficient in terms of network size as well as feed forward time yielding significantly high frame-rate reconstructions. At the same time, our simpler network achieves comparable accuracy in terms of surface and joint error w.r.t. majority of state-of-the-art techniques on three publicly available datasets. The proposed paradigm can theoretically learn local surface deformations induced by body shape variations which the PCA space of parametric body models can't capture. In addition to predicting the body model, we also show the generalizability of our proposed idea for solving a similar task with different structure - non-rigid hand mesh reconstructions from a monocular image.

\noindent To summarize, the key contributions of this work are:
\begin{itemize}
\item We propose a simple end-to-end multi-branch, multi-task deep network that exploits a "structured point cloud" to recover a smooth and fixed topology mesh model from a monocular image.
\item The proposed paradigm can theoretically learn local surface deformations induced by body shape variations which the PCA space of parametric body models can't capture.
\item The simplicity of the model makes it efficient in terms of network size as well as feed forward time yielding significantly high frame-rate reconstructions, while simultaneously achieving comparable accuracy in terms of surface and joint error, as shown on three publicly available datasets.
\item We also show the generalizability of our proposed paradigm for a similar task of reconstructing the hand mesh models from a monocular image.
\end{itemize}

\section{Related Work}
\label{sec:related}

\textbf{Estimating 3D Body Models:} The traditional approach for parametric body model fitting entails iteratively optimizing an objective function with 2D supervision in the form of silhouettes, 2D key points etc~\cite{sigal2008combined,guan2009estimating, keepSMPL2016,lassner2017unite}. However, they often involve manual intervention and are time-consuming to solve as well as susceptible to converge at local optima.

On the deep learning front, ~\cite{kanazawa2018end} proposes an iterative regression with 3D and 2D joint loss as a feedback and an adversarial supervision for each joint. However, this architecture has a large number of networks and hence is not easy to train. ~\cite{omran2018neural} predicts a colour-coded body segmentation that is used as a prior to predict the parameters. Similarly, in~\cite{pavlakos2018learning}, 2D heatmaps and silhouettes are predicted first, which are then used to predict the pose and shape parameters. All of the above methodologies calculate loss on 2D keypoints or silhouette projections of the rendered mesh, which significantly slows down training time (due to model complexity), in addition to requiring additional supervision. ~\cite{varol2018bodynet} proposes a complex multi-task network with total of six networks (having respective losses computed on 2D and 3D joint locations, 2D segmentation mask, volumetric grid and sillhoute reprojection of volumetric and SMPL model).
This makes it a significantly heavy network with a longer feed forward time. The focus of reconstruction is to retrieve the boundary of the subject in 3D space. However, in a volumetric representation, predicting the volume within the surface is counterproductive. On the other hand, we focus on direct image to mesh vertex regression for recovering the surface points. The most recent state-of-the-art work proposed in \cite{kolotouros2019cmr} also recovers sparse surface points using Graph Neural Network(GCN). However GCNs experience troubles learning the global structure because of its neighbourhood aggregation scheme \cite{xu2018how}.

\textbf{Estimating Hand Models:} While most of the hand recovery methods typically estimate the 3D pose from one or multiple RGB/Depth images, hand shape estimation hasn't been extensively explored. For a detailed survey of the field, we refer to~\cite{supancic2015depth,yuan2018depth}. 
Recent effort in~\cite{malik2018deephps} was the first attempt to predict both the pose and the vertex based full 3D mesh representation (surface shape) from a single depth image. The recently proposed MANO~\cite{romero2017embodied} model is an SMPL like model that describes both the shape and pose, and is learned from thousands of high resolution scans. ~\cite{boukhayma20193d} predicted the MANO parameters from a monocular RGB image, but, they don't show much shape variations. ~\cite{ge20193d} use a graph CNN to recover the hand surface from monocular RGB image of the hand. 

\section{Proposed Method: HumanMeshNet}
\label{sec:method}
In order to learn this structured point cloud, we use an encoder- and multi-decoder model, which we describe in this section.
Figure~\ref{fig:arch} gives an overview of our end-to-end pipeline.
Our model consists of three primary phases: \\

\noindent\textbf{Phase 1 - RGB to Partwise Segmentation: }Given an input RGB image of size 224x224, we first predict a discrete body part label for each pixel in the input image (for a total of 24 body parts) using just the body part labelling network from~\cite{alp2018densepose}. A part-wise segmentation enables a tracking of the human body in the image, making it easier for shape estimation.
\\

\noindent\textbf{Phase 2 - Image Encoders and Joint Embedding: } Both the RGB image and segmentation mask are passed through separate encoders, each a Resnet-18, and their respective CNN feature vectors, each of dimension 1000 are concatenated together to obtain a joint embedding.

\begin{figure}[h!]
\begin{subfigure}[b]{0.24\linewidth}
\includegraphics[width=0.75\linewidth]{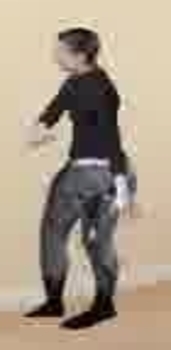}
\caption{}\label{fig:human}
\end{subfigure}
\begin{subfigure}[b]{0.24\linewidth}
\includegraphics[width=0.75\linewidth]{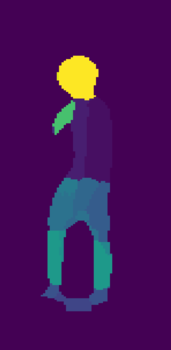}
\caption{}\label{fig:human}
\end{subfigure}
\begin{subfigure}[b]{0.24\linewidth}
\includegraphics[width=0.75\linewidth]{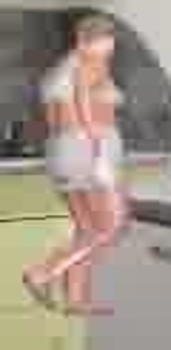}
\caption{}\label{fig:human}
\end{subfigure}
\begin{subfigure}[b]{0.24\linewidth}
\includegraphics[width=0.75\linewidth]{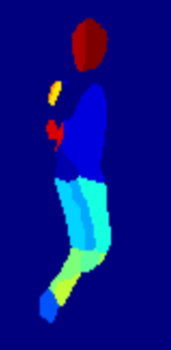}
\caption{}\label{fig:human}
\end{subfigure}
\caption{Noisy Segmentation Masks predicted from images (a) and (c) in Phase 1. The figure shows (b) missing body part masks (d) confusing between leg limbs.}
\label{fig:noisyseg}
\end{figure}

Such fusion of RGB and segmentation mask was employed to combine complementary information from each modality. This is important as segmentation mask predictions can be very noisy in many scenarios (see Figure~\ref{fig:noisyseg}), e.g., low lighting, distance of the person from the camera, sensing noise etc., leading to failures like interchanged limbs or missing limbs. \\

\noindent\textbf{Phase 3 - Multi-branch Predictions:} From our concatenated feature embedding, we branch out into two complimentary tasks via Fully Connected layers (FCs). Each branch consists of two FCs, each of dimension 1000 followed by the respective output dimensions for the 3D joints, and 3D surface respectively. It is to be noted that our predictions are in camera frame.

\noindent\textbf{Loss Function.} We use a multi-branch loss functions to train our network i.e, $L_S$, $L_J$ and $L_{JS}$. We regularized the loss functions such that they contribute equally to the overall loss. This translates to Equation~\ref{eqn:L}.
\begin{equation} \label{eqn:L}
L = L_S + (\lambda_1*L_J) + (\lambda_2*L_{JS})
\end{equation}
The surface loss $L_S$ in Equation~\ref{eqn:L_S} gives the vertex-wise Euclidean distance between the predicted vertices $V_i$ and ground truth vertices $\hat{V_i}$ for the 3D mesh prediction branch in Figure~\ref{fig:arch} (e).
\begin{equation} \label{eqn:L_S}
L_S = \sum_{\forall V_i} ||V_i - \hat{V_i}||_2
\end{equation}
However, this loss does not ensure prediction of smooth surfaces as each vertex is independently predicted.

Nevertheless, each mesh vertex has a neighborhood structure that can be used to further refine the estimate of individual vertex.
Here we make use of smoothing regularisation~\cite{sorkine2004laplacian} (as shown in Equation~\ref{eqn:1}), where the position of each vertex, $V_i$, is replaced by the average position, of its neighbours $N(V_j)$.
\begin{equation} \label{eqn:1}
V_i = \frac{1}{|N(V_i)|}\sum_{V_j \in N(V_i)}V_j \;\;\;\;\;\; \forall V_i
\end{equation}
This is achieved by first applying the smoothness mesh regularization given by Equation~\ref{eqn:1} and then calculating $L_S$. This helps in limiting the number of surface jitters or irregularities.

In order to enforce 3D joints consistency, we minimize joint loss $L_J$ defined in Equation~\ref{eqn:L_J}, which gives the euclidean distance between the predicted joints $J_i$ and ground truth joints $\hat{J_i}$ in the 3D joint prediction branch as shown in Figure~\ref{fig:arch}(d).
\begin{equation} \label{eqn:L_J}
L_J = \sum_{\forall J_i} ||J_i - \hat{J_i}||_2
\end{equation}

The 3D joints $JS_i$ under the surface are recovered using the SMPL joint regressor~\cite{SMPL:2015}.  We also minimize the loss $L_{JS}$ defined in Equation~\ref{eqn:L_JS} which gives the euclidean distance between the joints $J_i$ predicted from the joints branch and the joints $JS_i$ from the surface branch. It helps both the branches to learn consistently with each other. \\
\begin{equation} \label{eqn:L_JS}
L_{JS} = \sum_{\forall J_i} ||J_i - JS_i||_2
\end{equation}

\noindent {\bf Network Variants:} We define two different variants of HumanMeshNet in order to perform an extensive analysis:
\begin{enumerate}
  \item [(a)] HumanMeshNet (HMNet) - The base version which uses an "off-the-shelf" body part segmentation network (~\cite{alp2018densepose}).
  \item [(b)] HumanMeshNetOracle (HMNetOracle) - A refined version using a more accurate body part segmentation given by the dataset. However, in some datasets (e.g., UP-3D,~\cite{lassner2017unite}), these segmentation masks can be noisy due to manual annotations.
\end{enumerate}

\section{Experiments \& Results}
\label{sec:exp}
In this section, we show a comprehensive evaluation of the proposed model and benchmark against the state-of-the-art optimization and deep learning based Parametric (P), Volumetric (V) and Surface based (S) reconstruction algorithms.  It is to be noted that we train on each dataset separately and report on its given test sets. All of the trained models and code shall be made publicly available, along with a working demo. Please view our supplementary video for more results.


\subsection{Datasets}
\label{sec:datasets}
\noindent \textbf{SURREAL~\cite{varol2017learning}}: This dataset  provides synthetic image examples with 3D shape ground truth. The dataset draws poses from MoCap~\cite{ionescu2014human3} and body shapes from body scans~\cite{robinette2002civilian} to generate valid SMPL instances for each image. Although this dataset is synthetically generated, it emulates complex real poses and shapes, coupled with challenging input images that contain background clutter and are reflective with low resolution. It has total 1.6 million training and 15,000 test samples.\\

\begin{figure}[h!]
\begin{subfigure}[b]{0.16\linewidth}
\includegraphics[width=\linewidth]{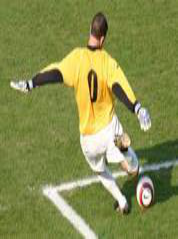}
\caption*{(a)}
\end{subfigure}
\begin{subfigure}[b]{0.16\linewidth}
\includegraphics[width=\linewidth]{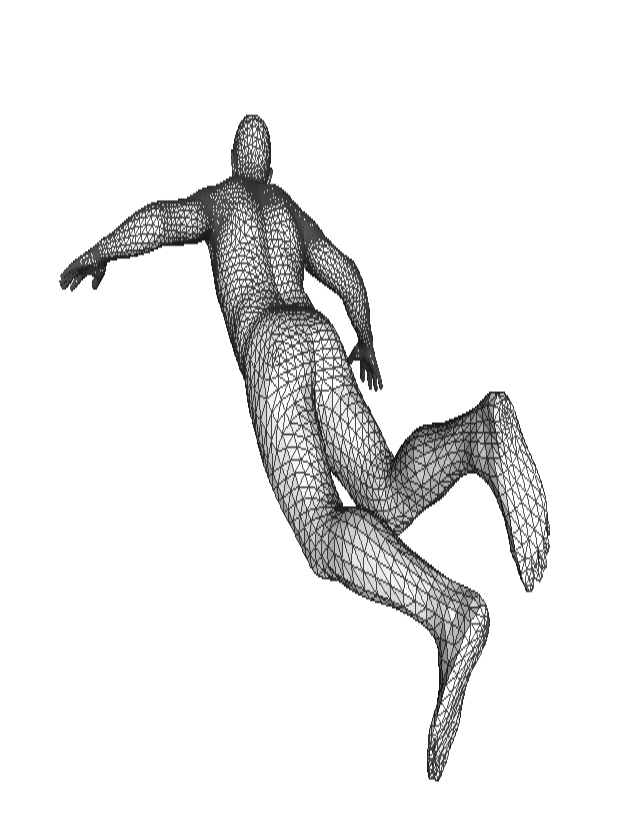}
\caption*{(b)}
\end{subfigure}
\begin{subfigure}[b]{0.16\linewidth}
\includegraphics[width=\linewidth]{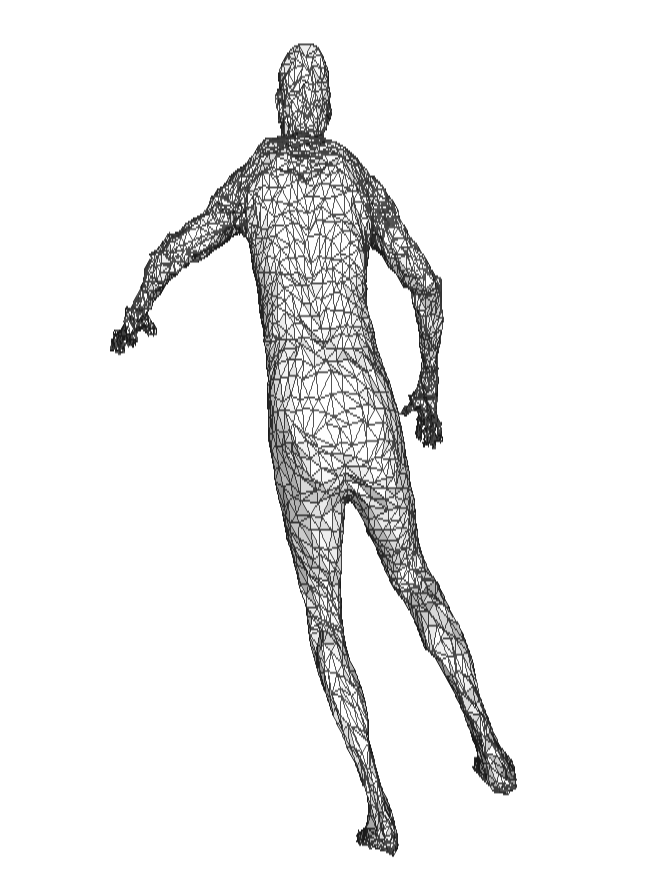}
\caption*{(c)}
\end{subfigure}
\begin{subfigure}[b]{0.15\linewidth}
\includegraphics[width=0.9\linewidth]{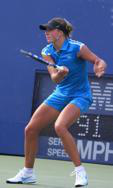}
\caption*{(a)}
\end{subfigure}
\begin{subfigure}[b]{0.15\linewidth}
\includegraphics[width=0.9\linewidth]{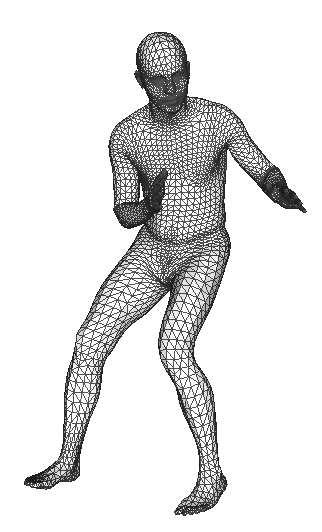}
\caption*{(b)}
\end{subfigure}
\begin{subfigure}[b]{0.15\linewidth}
\includegraphics[width=0.9\linewidth]{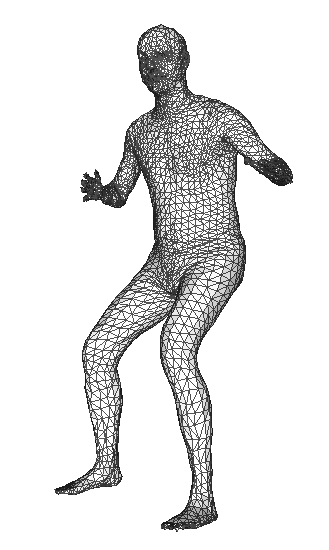}
\caption*{(c)}
\end{subfigure}
\caption{This figure depicts the quality of ground truth fits provided on UP-3D. (a) The input RGB image is fit using SMPLify~\cite{keepSMPL2016} to give (b) the ground truth. Our fit (c) makes use of more accurate markers or keypoints in a multi-branch setup, to account for noisy ground truth mesh data.}
\label{fig:up3d_faults}
\end{figure}

\noindent \textbf{UP-3D~\cite{lassner2017unite}}: It is a recent dataset that collects color images from 2D human pose benchmarks and uses an extended version of SMPLify~\cite{keepSMPL2016} to provide 3D human shape candidates. The candidates were evaluated by human annotators to select only the images with good 3D shape fits. It comprises of 8,515 images, where 7,126 are used for training and 1,389 for testing. However, the ground truth meshes are sometimes inaccurately generated as shown in~Figure~\ref{fig:up3d_faults}.
We separately train the network and report results on full test set of UP-3D. \\

\noindent \textbf{Human3.6M~\cite{ionescu2014human3}}: It is a large-scale pose dataset that contains multiple subjects performing typical actions like "eating" and "walking" in a lab environment. It consists of a downsampled version of the original data with 300,000 image-3D joint pairs for training and 100,000 such for testing. Since ground truth 3D meshes for any of the commonly reported protocols~\cite{bogo2016keep} for evaluation aren't available anymore, we finetune SURREAL-pretrained network using joint loss only. We report the joint reconstruction error (trained as per Protocol 2 of ~\cite{bogo2016keep}) and therefore compare with those methods that don't use mesh supervision for this dataset in Table~\ref{table:h36m}. \\

\begin{figure*}[h!]
\begin{subfigure}[b]{0.075\linewidth}
\includegraphics[width=\linewidth]{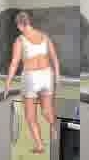}
\end{subfigure}
\begin{subfigure}[b]{0.075\linewidth}
\includegraphics[width=\linewidth]{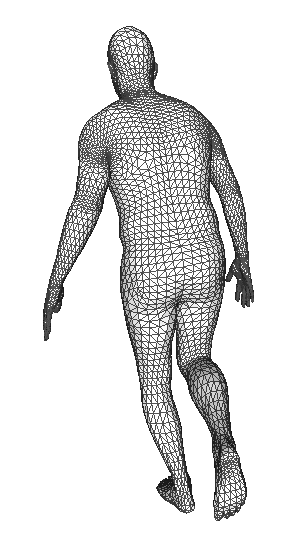}
\end{subfigure}
\begin{subfigure}[b]{0.075\linewidth}
\includegraphics[width=\linewidth]{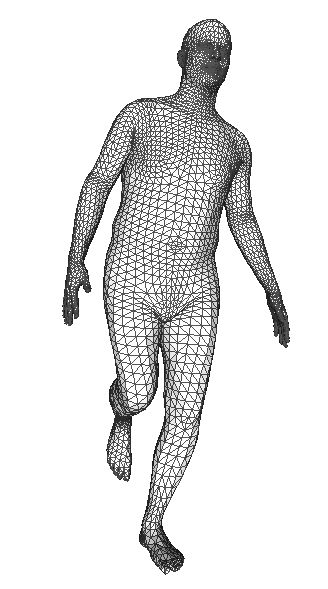}
\end{subfigure}
\begin{subfigure}[b]{0.075\linewidth}
\includegraphics[width=\linewidth]{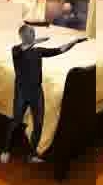}
\end{subfigure}
\begin{subfigure}[b]{0.075\linewidth}
\includegraphics[width=\linewidth]{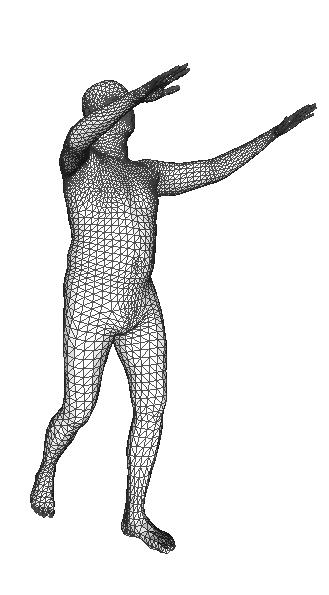}
\end{subfigure}
\begin{subfigure}[b]{0.075\linewidth}
\includegraphics[width=\linewidth]{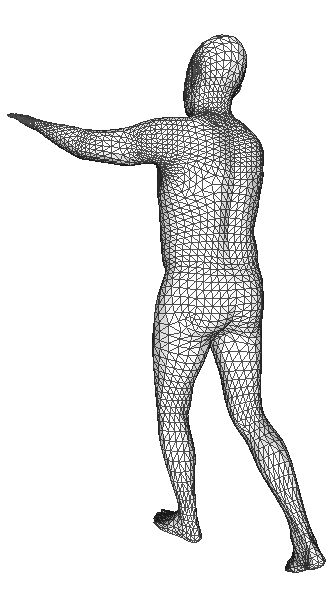}
\end{subfigure}\;\;\;\;
\begin{subfigure}[b]{0.075\linewidth}
\includegraphics[width=\linewidth]{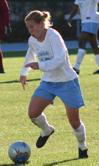}
\end{subfigure}
\begin{subfigure}[b]{0.075\linewidth}
\includegraphics[width=\linewidth]{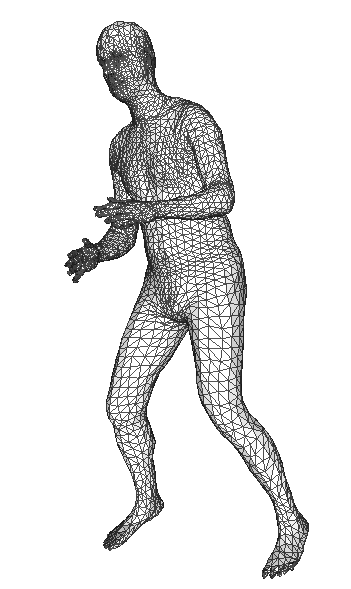}
\end{subfigure}
\begin{subfigure}[b]{0.075\linewidth}
\includegraphics[width=\linewidth]{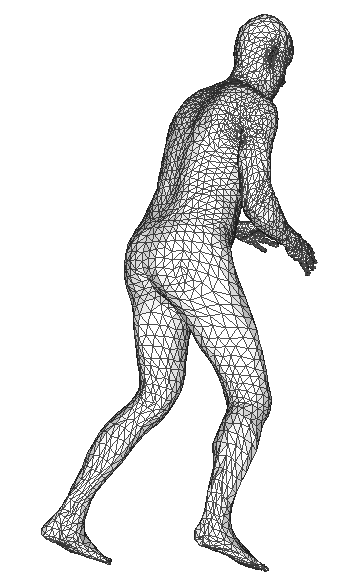}
\end{subfigure}
\begin{subfigure}[b]{0.075\linewidth}
\includegraphics[width=\linewidth]{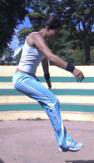}
\end{subfigure}
\begin{subfigure}[b]{0.075\linewidth}
\includegraphics[width=\linewidth]{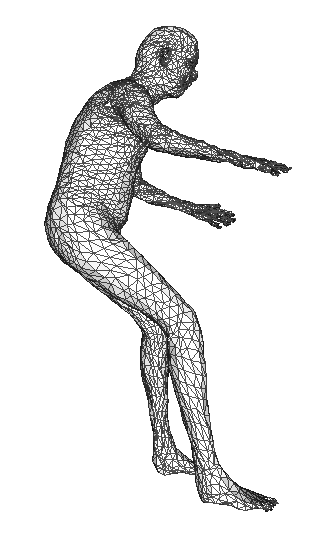}
\end{subfigure}
\begin{subfigure}[b]{0.075\linewidth}
\includegraphics[width=\linewidth]{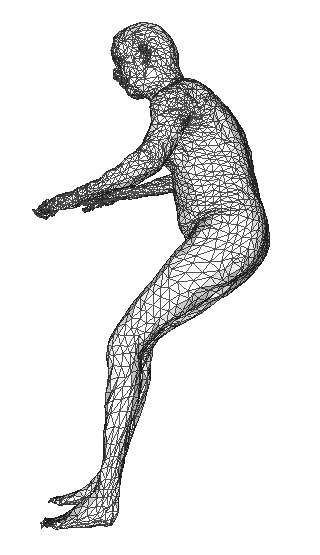}
\end{subfigure}
\begin{subfigure}[b]{0.075\linewidth}
\includegraphics[width=\linewidth]{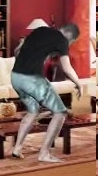}
\end{subfigure}
\begin{subfigure}[b]{0.075\linewidth}
\includegraphics[width=\linewidth]{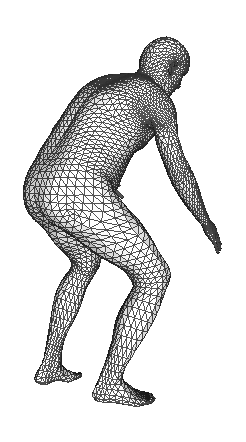}
\end{subfigure}
\begin{subfigure}[b]{0.075\linewidth}
\includegraphics[width=\linewidth]{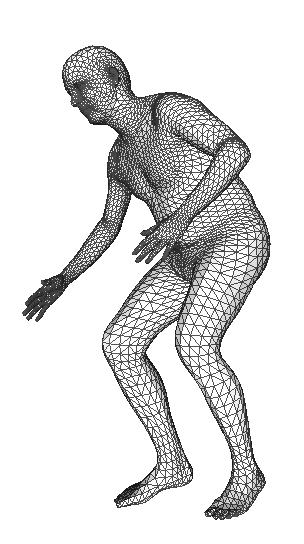}
\end{subfigure}
\begin{subfigure}[b]{0.075\linewidth}
\includegraphics[width=\linewidth]{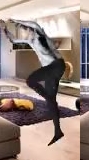}
\end{subfigure}
\begin{subfigure}[b]{0.075\linewidth}
\includegraphics[width=\linewidth]{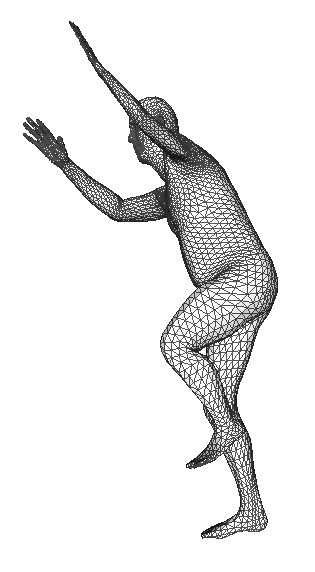}
\end{subfigure}
\begin{subfigure}[b]{0.075\linewidth}
\includegraphics[width=\linewidth]{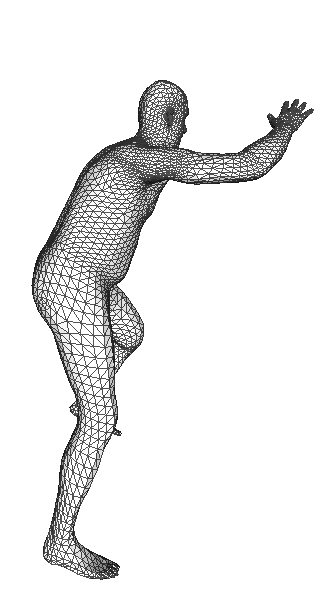}
\end{subfigure}\;\;\;\;
\begin{subfigure}[b]{0.075\linewidth}
\includegraphics[width=\linewidth]{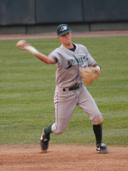}
\end{subfigure}
\begin{subfigure}[b]{0.075\linewidth}
\includegraphics[width=\linewidth]{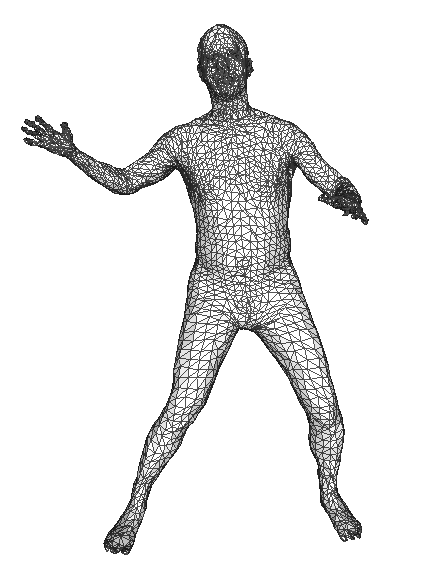}
\end{subfigure}
\begin{subfigure}[b]{0.075\linewidth}
\includegraphics[width=\linewidth]{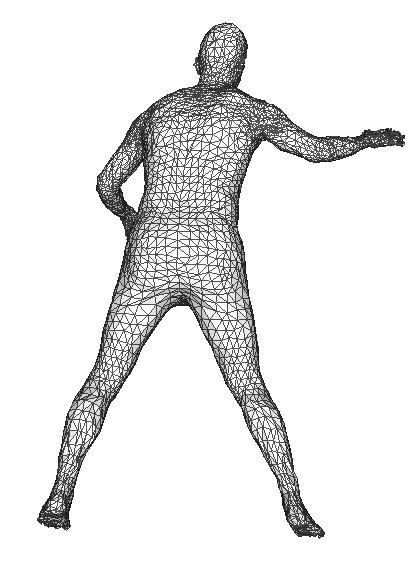}
\end{subfigure}
\begin{subfigure}[b]{0.075\linewidth}
\includegraphics[width=\linewidth]{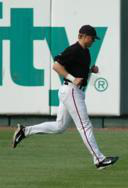}
\end{subfigure}
\begin{subfigure}[b]{0.075\linewidth}
\includegraphics[width=\linewidth]{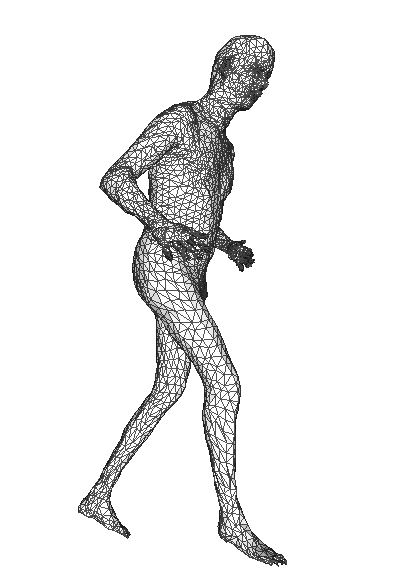}
\end{subfigure}
\begin{subfigure}[b]{0.075\linewidth}
\includegraphics[width=\linewidth]{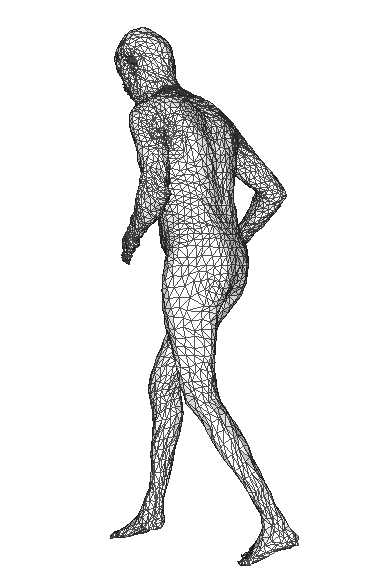}
\end{subfigure}
\begin{subfigure}[b]{0.075\linewidth}
\includegraphics[width=\linewidth]{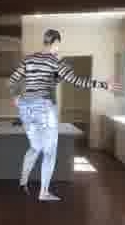}
\caption*{(a)}
\end{subfigure}
\begin{subfigure}[b]{0.075\linewidth}
\includegraphics[width=\linewidth]{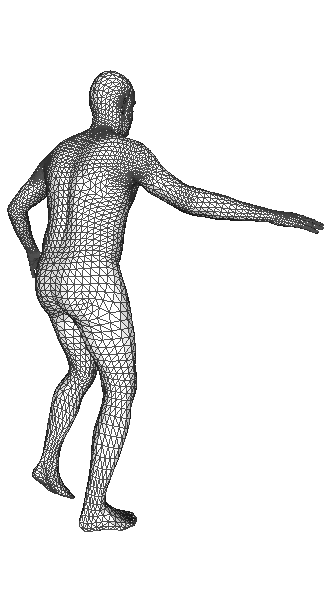}
\caption*{(b)}
\end{subfigure}
\begin{subfigure}[b]{0.075\linewidth}
\includegraphics[width=\linewidth]{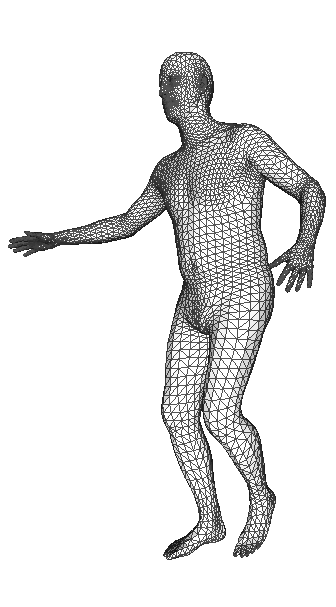}
\caption*{(c)}
\end{subfigure}\;\;\;
\begin{subfigure}[b]{0.075\linewidth}
\includegraphics[width=\linewidth]{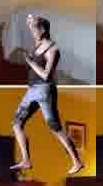}
\caption*{(a)}
\end{subfigure}
\begin{subfigure}[b]{0.075\linewidth}
\includegraphics[width=\linewidth]{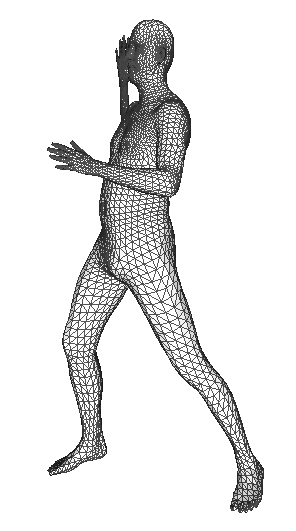}
\caption*{(b)}
\end{subfigure}
\begin{subfigure}[b]{0.075\linewidth}
\includegraphics[width=\linewidth]{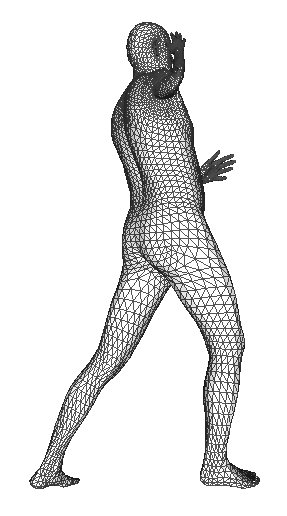}
\caption*{(c)}
\end{subfigure}\;\;\;\;\;
\begin{subfigure}[b]{0.075\linewidth}
\includegraphics[width=\linewidth]{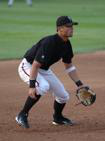}
\caption*{(a)}
\end{subfigure}
\begin{subfigure}[b]{0.075\linewidth}
\includegraphics[width=\linewidth]{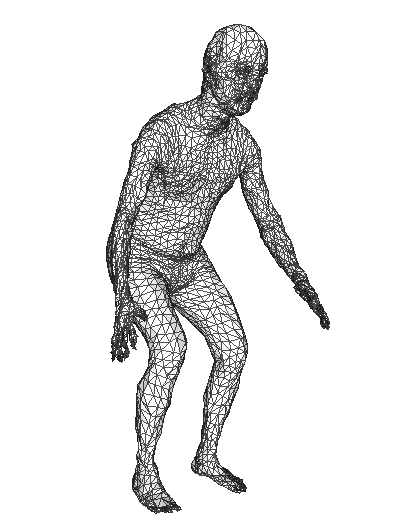}
\caption*{(b)}
\end{subfigure}
\begin{subfigure}[b]{0.075\linewidth}
\includegraphics[width=\linewidth]{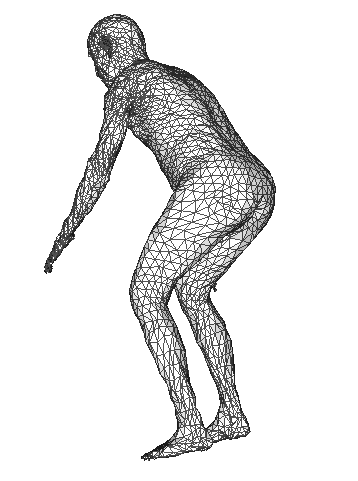}
\caption*{(c)}
\end{subfigure}\;\;
\begin{subfigure}[b]{0.075\linewidth}
\includegraphics[width=\linewidth]{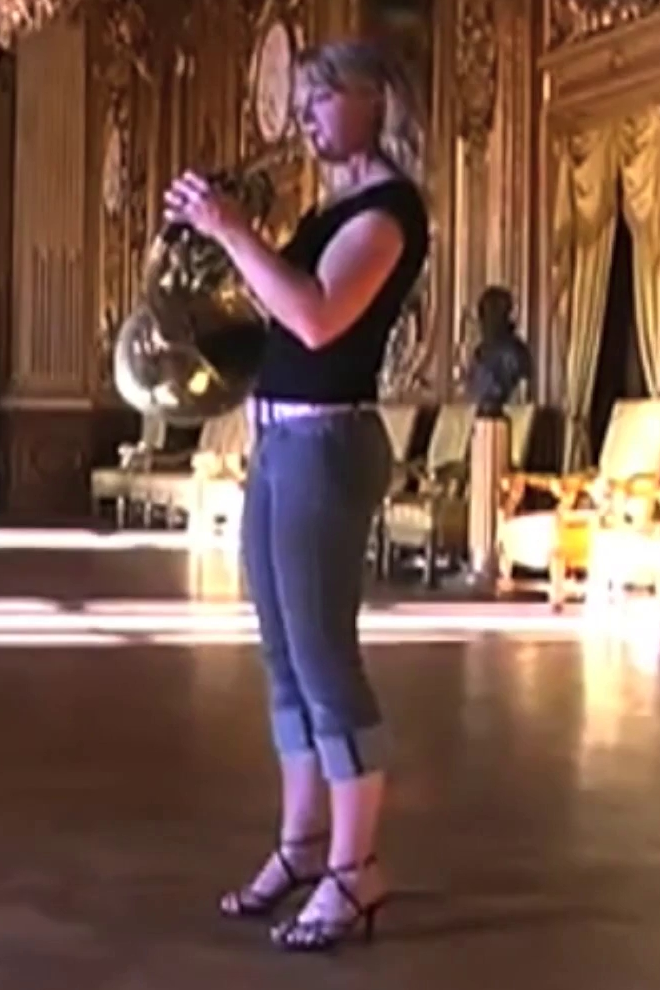}
\caption*{(a)}
\end{subfigure}
\begin{subfigure}[b]{0.075\linewidth}
\includegraphics[width=\linewidth]{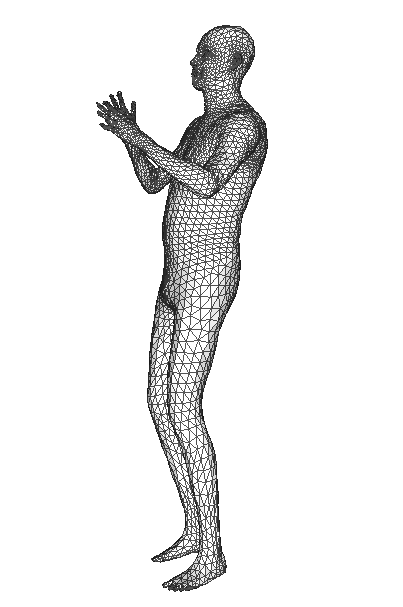}
\caption*{(b)}
\end{subfigure}
\begin{subfigure}[b]{0.075\linewidth}
\includegraphics[width=\linewidth]{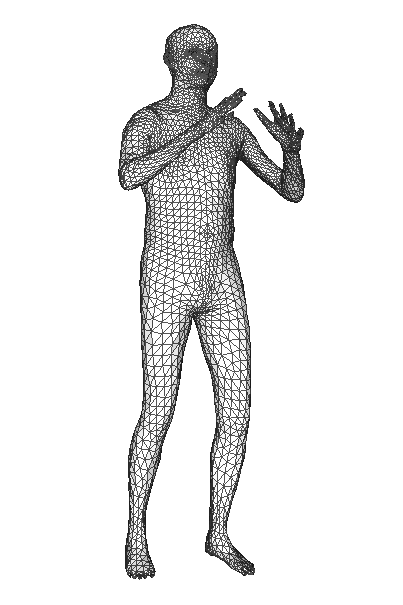}
\caption*{(c)}
\end{subfigure}
\caption{Qualitative Results on SURREAL~\cite{varol2017learning} (first six columns) and UP-3D~\cite{lassner2017unite} (next six columns) where (a) represents the input view, (b) our mesh reconstruction aligned to the input view, and (c) aligned to another arbitrary view.}
\label{fig:qual_results}
\end{figure*}

\begin{table*}
\centering
\begin{tabular}{| c | c | c c c c |}
  \hline
    &  & Surface & Joints & PA. Surface & PA. Joint\\
   Output & Method & Error & Error & Error & Error\\
   \hline
  P & Pavlakos \textit{et al.}~\cite{pavlakos2018learning} & 117.7 & - & - & - \\
  P & Lasner \textit{et al.}~\cite{lassner2017unite} & 169.8 & - & - & -\\
  P  & NBF~\cite{omran2018neural} & - & - & - & 82.3\\
  \hline
  V & BodyNet~\cite{varol2018bodynet} & 80.1 & - & - & - \\
  \hline
  S & Baseline & 151.4 & 130.8 & 93.8 & 83.7\\
  S & HMNet & 130.4 & 112.5 & 77.6 & 69.6\\
  S & HMNetOracle & \textbf{60.3} & \textbf{51.5} & \textbf{42.9} & \textbf{37.9} \\
  \hline
\end{tabular}
\caption{Comparison with other methods on UP3D's full test set~\cite{lassner2017unite} }
\label{table:up3d}
\end{table*}
%

\subsection{Implementation Details}

\noindent\textbf{Data Pre-processing} We use the ground truth bounding boxes from each of the datasets to obtain a square crop of the human. This is a standard step performed by most comparative 3D human reconstruction models. \\

\noindent \textbf{Network Training }We use Nvidia's GTX 1080Ti, with 11GB of VRAM to train our models. A batch size of 64 is used for SURREAL and Human 3.6M datasets, and a batch size of 16 for UP3D dataset. We use the ADAM optimizer having an initial learning rate of $10^{-4}$, to get optimal performance. Attaining convergence on the SURREAL and Human3.6M takes 18 hours each, while on UP-3D takes 6 hours. We use the standard splits given by the datasets, for benchmarking, as indicated in Section~\ref{sec:datasets}.\\
\begin{figure*}

\begin{subfigure}[b]{0.105\linewidth}
\includegraphics[width=\linewidth]{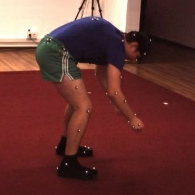}
\end{subfigure}
\begin{subfigure}[b]{0.105\linewidth}
\includegraphics[width=\linewidth]{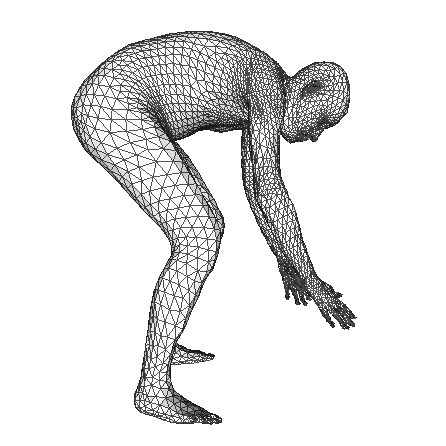}
\end{subfigure}
\begin{subfigure}[b]{0.105\linewidth}
\includegraphics[width=\linewidth]{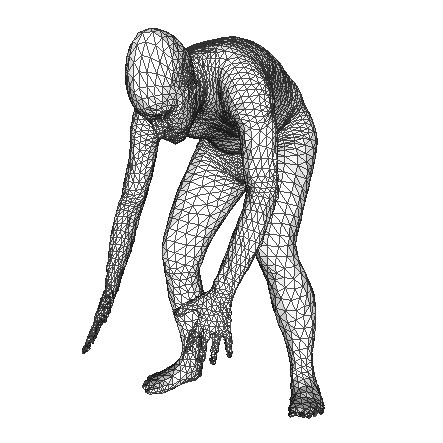}
\end{subfigure}
\begin{subfigure}[b]{0.105\linewidth}
\includegraphics[width=\linewidth]{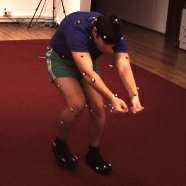}
\end{subfigure}
\begin{subfigure}[b]{0.105\linewidth}
\includegraphics[width=\linewidth]{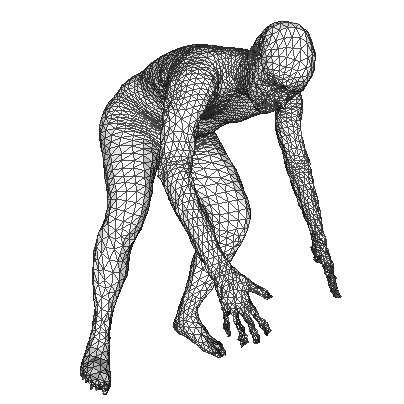}
\end{subfigure}
\begin{subfigure}[b]{0.105\linewidth}
\includegraphics[width=\linewidth]{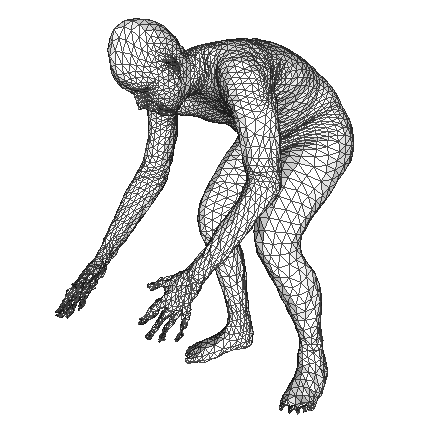}
\end{subfigure}
\begin{subfigure}[b]{0.105\linewidth}
\includegraphics[width=\linewidth]{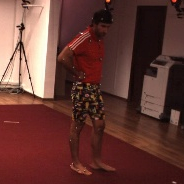}
\end{subfigure}
\begin{subfigure}[b]{0.105\linewidth}
\includegraphics[width=\linewidth]{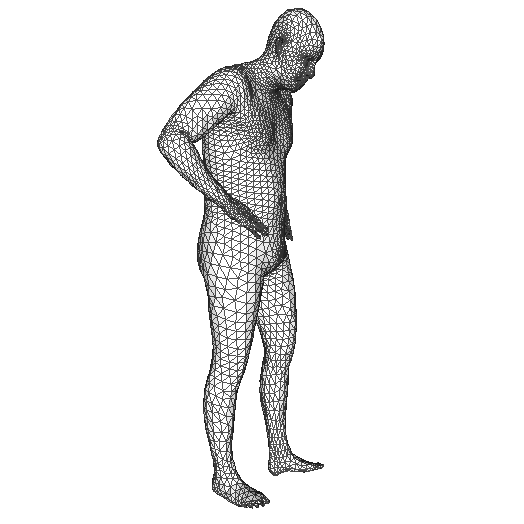}
\end{subfigure}
\begin{subfigure}[b]{0.105\linewidth}
\includegraphics[width=\linewidth]{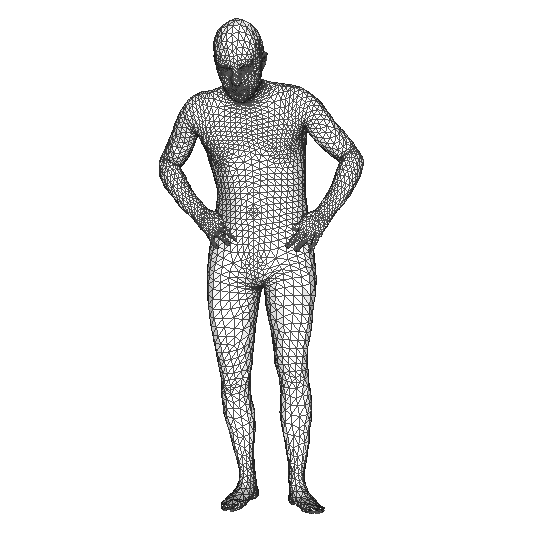}
\end{subfigure}

\begin{subfigure}[b]{0.105\linewidth}
\includegraphics[width=\linewidth]{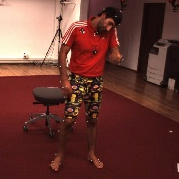}
\caption*{(a)}
\end{subfigure}
\begin{subfigure}[b]{0.105\linewidth}
\includegraphics[width=\linewidth]{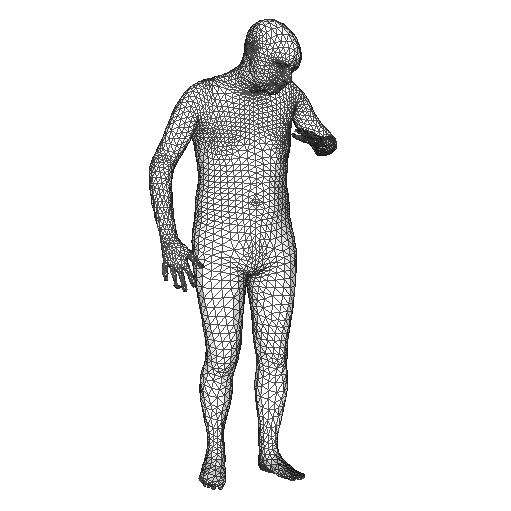}
\caption*{(b)}
\end{subfigure}
\begin{subfigure}[b]{0.105\linewidth}
\includegraphics[width=\linewidth]{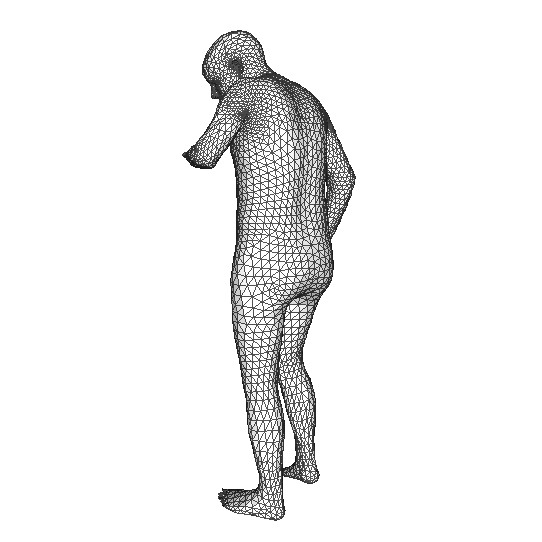}
\caption*{(c)}
\end{subfigure}
\begin{subfigure}[b]{0.105\linewidth}
\includegraphics[width=\linewidth]{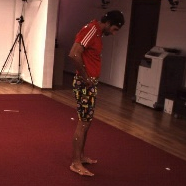}
\caption*{(a)}
\end{subfigure}
\begin{subfigure}[b]{0.105\linewidth}
\includegraphics[width=\linewidth]{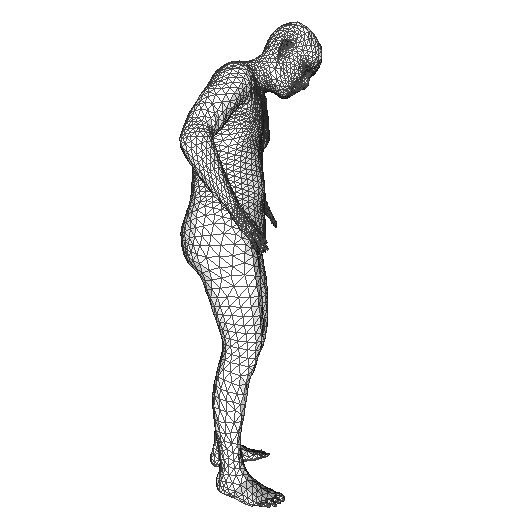}
\caption*{(b)}
\end{subfigure}
\begin{subfigure}[b]{0.105\linewidth}
\includegraphics[width=\linewidth]{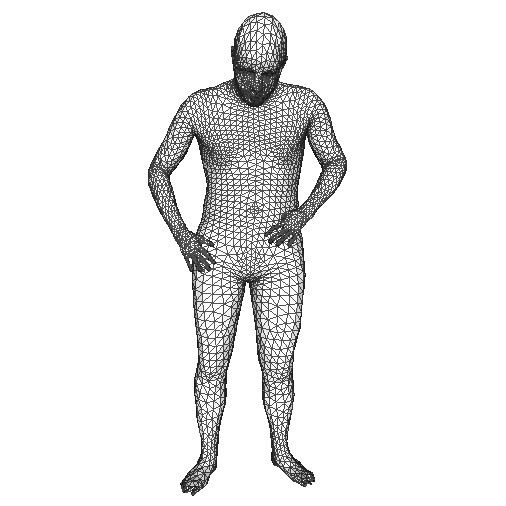}
\caption*{(c)}
\end{subfigure}
\begin{subfigure}[b]{0.105\linewidth}
\includegraphics[width=\linewidth]{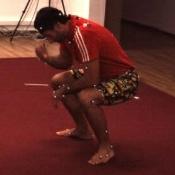}
\caption*{(a)}
\end{subfigure}
\begin{subfigure}[b]{0.105\linewidth}
\includegraphics[width=\linewidth]{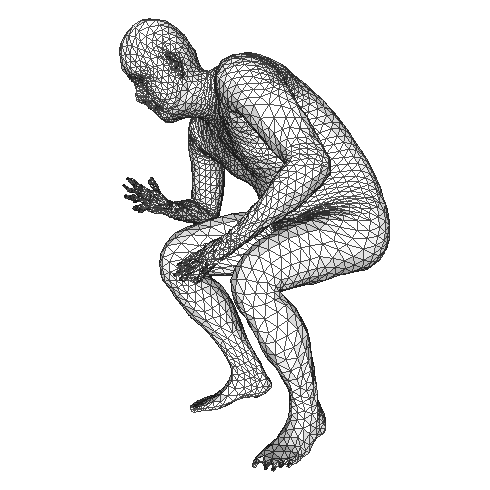}
\caption*{(b)}
\end{subfigure}
\begin{subfigure}[b]{0.105\linewidth}
\includegraphics[width=\linewidth]{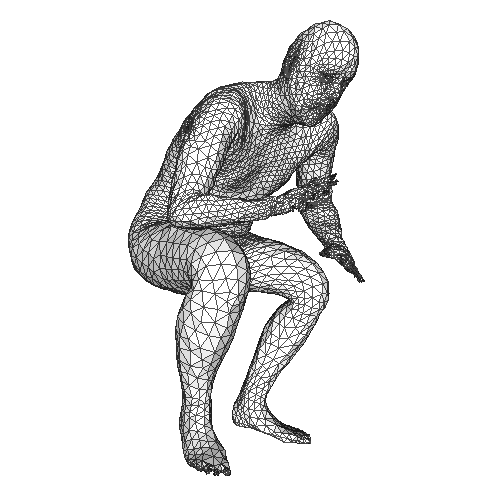}
\caption*{(c)}
\end{subfigure}
\caption{Qualitative Results on Human3.6M, where (a) represents the input view, (b) our mesh reconstruction aligned to the input view, and (c) aligned to another arbitrary view.}
\label{fig:qual_results_h36m}
\end{figure*}

\noindent{\textbf{Procrustes Analysis (PA)}} In order to evaluate the quality of the reconstructed mesh, we also report results after solving the Orthogonal Procrustes problem~\cite{robinette2002civilian}, in which we scale the output to the size of the ground truth and solve for rotation. Additionally, we also quantitatively evaluate without this alignment.  \\

\noindent \textbf{Evaluation Metric}
\begin{enumerate}
  \item [(a)] Surface Error (mm): Gives the mean-per-vertex error between the ground truth and predicted mesh.
  \item [(b)] Joint Error (mm): Gives the mean-per joint error between the ground truth and predicted joints. All reported results are obtained from the underlying joints of the mesh, rather than the alternate branch, unless otherwise mentioned.
  \item [(c)] PA. Surface/Joint Error (mm): It is the surface/joint error after Procrustes Analysis (PA).
\end{enumerate}

\subsection{Comparison with State-of-the-art}

\noindent \textbf{Baseline }We define our baseline as the direct prediction of a point cloud from an RGB image, using a Resnet-50. This enables us to show the novelty introduced by our pipeline and the usefulness of learning in this output space.\\

\begin{table}[h!]
\begin{center}
\begin{tabular}{| c | c | c c |}
   \hline
    &  & Surface & Joint \\
  Output  & Method  & Error & Error \\
   \hline
  \multirow{3}{*}{P} & Tung \textit{et al.}~\cite{tung2017self} & 74.5 & 64.4 \\
  & Pavlakos \textit{et al.}~\cite{pavlakos2018learning} & 151.5 & - \\
  & SMPLR ~\cite{madadi2018smplr}                       & 75.4 & 55.8 \\
  \hline
  V & BodyNet~\cite{varol2018bodynet} & 65.8 & -\\
  \hline
  \multirow{2}{*}{S} & Baseline & 101 & 85.7\\
   & HMNet[subsampled] & 86.9	& 72.4 \\
   & HMNet & 86.6 & 71.9 \\
   & HMNetOracle & \textbf{63.5} & \textbf{49.1} \\
  \hline
\end{tabular}
\caption{Comparison with state-of-the-art methods on SURREAL's test set~\cite{varol2017learning}}
\label{table:surreal}
\end{center}
\end{table}

\begin{table}[h!]
\begin{center}
\begin{tabular}{| c | c | c |}
\hline
3D mesh &  & PA. Joint \\
Supervision & Method & Error \\
\hline
\multirow{7}{*}{No} 
& Ramakrishnan \textit{et al.}~\cite{ramakrishna2012reconstructing}   & 157.3\\
& Zhou \textit{et al.}~\cite{zhou20153d}                              & 106.7\\
& SMPLify ~\cite{bogo2016keep}                                        & 82.3\\
& SMPLify 91 kps ~\cite{lassner2017unite}                             & 80.7\\

& Pavlakos \textit{et al.}~\cite{pavlakos2018learning}                & 75.9\\
& HMR ~\cite{kanazawa2018end}                                         & \textbf{56.8}\\
& HMNet(Ours)                                                         & 60.9\\
\hline
\multirow{3}{*}{Yes} & NBF ~\cite{omran2018neural}                                         & 59.9\\
& SMPLR ~\cite{madadi2018smplr}                                       & 56.4\\
& CMR ~\cite{kolotouros2019cmr}                                       & \textbf{50.1}\\
  \hline
\end{tabular}
\caption{Joint Reconstruction error as per Protocol 2 of Bogo \textit{et al.}~\cite{bogo2016keep} on Human 3.6M~\cite{ionescu2014human3}. Refer to Section~\ref{sec:datasets} for details on 3D mesh supervision}
\label{table:h36m}
\end{center}
\end{table}

\noindent \textbf{Results \& Discussion }  For qualitative results on all of the three datasets, refer to Figures~\ref{fig:qual_results}, ~\ref{fig:qual_results_h36m}. A large amount of training data is required to learn a vast range of poses and shapes. However, ~\cite{varol2017learning, pavlakos2018learning} show a good domain transfer to real data by training on the synthetic SURREAL dataset. Since our supervision is dominated by surface meshes, SURREAL plays an important role in benchmarking our method. We show comparable performance on it, as indicated by Table~\ref{table:surreal}. In Table~\ref{table:surreal}, we also show our results with a subsampled mesh (subsampled as per~\cite{kolotouros2019cmr}) from 6890 to 1723 vertices with almost no change in reconstruction error. This is a good proof of our hypothesis that there are far fewer representative points to learn in this structured point cloud. \\

UP-3D is an "in the wild" dataset, however has inaccurate ground truth mesh annotations, as shown in Figure~\ref{fig:up3d_faults}. Most circumvent this issue, by avoiding 3D supervision altogether and projecting back to a silhouette or keypoints~\cite{kanazawa2018end,pavlakos2018learning}. Further, training on such a small dataset doesn't provide a good generalisation. Therefore, we observe a higher error in HMNet. However, HMNetOracle produces a significant increase in accuracy with the increase in quality of the input image and segmentation mask (Table~\ref{table:ablation}). Similar to state of the art methods ~\cite{varol2018bodynet,venkat2018deep,kolotouros2019cmr}, we rely on 3D body supervision and providing more supervision like sillhoute and 2D keypoint loss like ~\cite{varol2018bodynet,kanazawa2018end} can improve the performance further. For Human3.6m, we compare with those that don't use mesh supervision (since this data is currently unavailable) and achieve comparable performance.

\subsection{Discussion}

\begin{figure}[h!]
\begin{subfigure}[b]{0.11\linewidth}
\includegraphics[width=\linewidth]{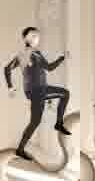}
\caption*{(a)}
\end{subfigure}
\begin{subfigure}[b]{0.11\linewidth}
\includegraphics[width=\linewidth]{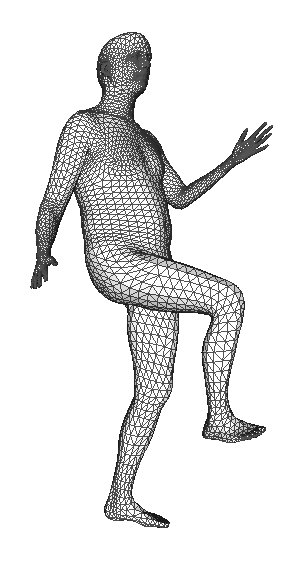}
\caption*{(b)}
\end{subfigure}
\begin{subfigure}[b]{0.11\linewidth}
\includegraphics[width=\linewidth]{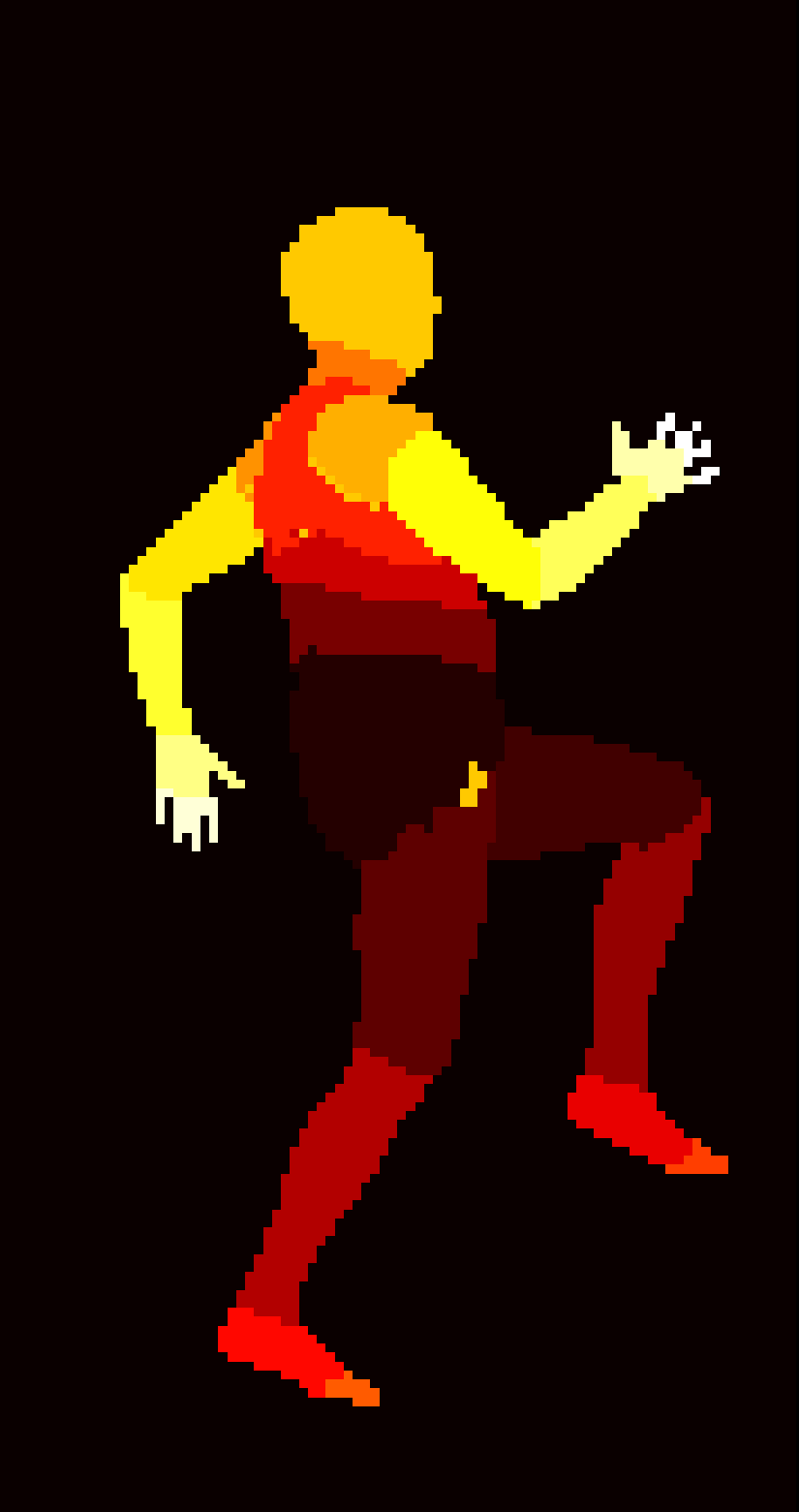}
\caption*{(c)}
\end{subfigure}
\begin{subfigure}[b]{0.11\linewidth}
\includegraphics[width=\linewidth]{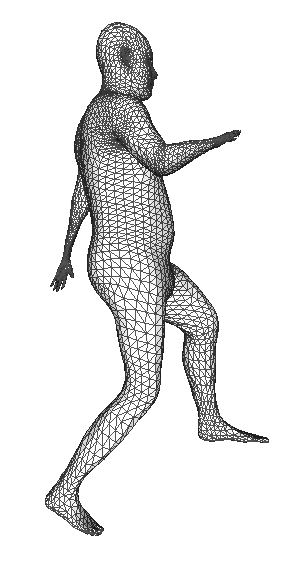}
\caption*{(d)}
\end{subfigure}
\begin{subfigure}[b]{0.11\linewidth}
\includegraphics[width=\linewidth]{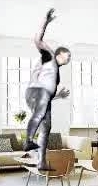}
\caption*{(a)}
\end{subfigure}
\begin{subfigure}[b]{0.11\linewidth}
\includegraphics[width=\linewidth]{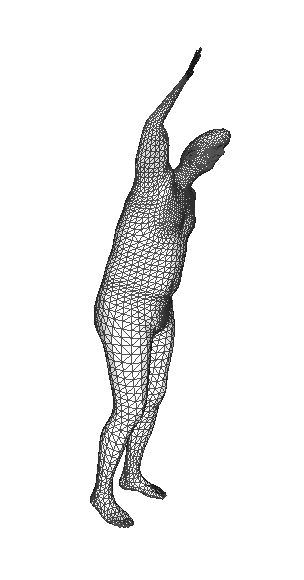}
\caption*{(b)}
\end{subfigure}
\begin{subfigure}[b]{0.11\linewidth}
\includegraphics[width=\linewidth]{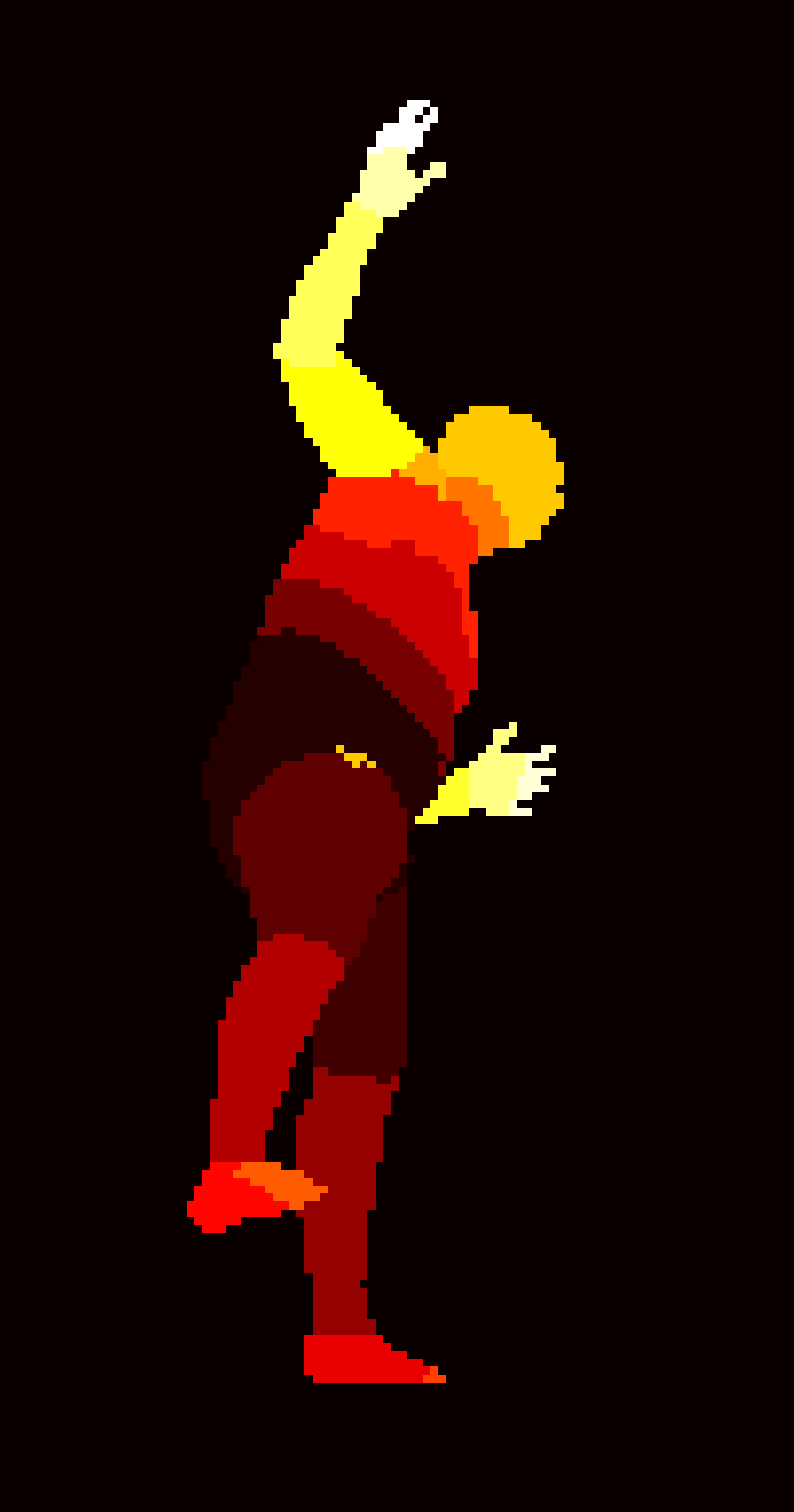}
\caption*{(c)}
\end{subfigure}
\begin{subfigure}[b]{0.11\linewidth}
\includegraphics[width=\linewidth]{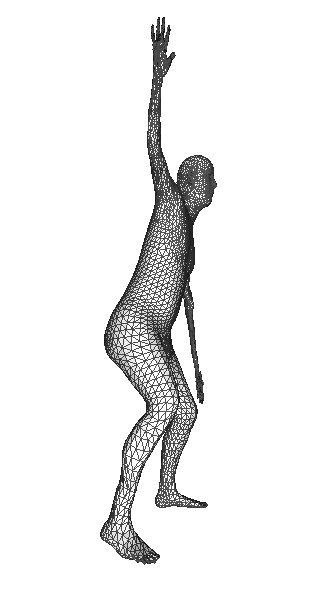}
\caption*{(d)}
\end{subfigure}
\caption{Given an input RGB image (a), this figure depicts a comparison of the baseline (b), against our output, HMNet (d). The predicted part-wise segmentation mask (c) assits HMNet to track the body parts and therefore solve the confusion between the legs as well as complex poses.
}
\label{fig:surreal_BL_comp}
\end{figure}

\noindent \textbf{Ablation Study: } Directly regressing the mesh from RGB leads to sub-par performance. Limbs are typically the origin of maximum error in reconstruction, and the segmentation mask provides a better tracking in scenarios such as leg-swap shown in Figure~\ref{fig:surreal_BL_comp}. The first two rows of Table~\ref{table:ablation} quantitatively explain this behaviour. Further, by having a more accurate segmentation mask, HMNetOracle achieves a significant reduction in surface error ($\downarrow 34.7mm$). In scenarios with inaccurate ground truth 3D (Figure~\ref{fig:up3d_faults}), the regularisation 3D joint loss in our multi-branch setup helps us in recovering better fits (row 4 for UP3D). In datasets such as Human3.6m where accurate MoCap markers are given, this multi-branch loss provides a good boost - with and without joint loss, the joint reconstruction error is 60.9mm v/s 67.3mm respectively.
\\

\begin{table}[h!]
\centering
\begin{tabular}{| c | c | c c |}
  \hline
  Config.& Input & PA. Surface & PA. Joint\\
         &    & Error & Error \\
   \hline
  Baseline & RGB & 93.8 & 83.7 \\
  Single Task & SM$_{DP}$ & 82.9 & 74.6\\
  Single Task & RGB+SM$_{DP}$ & 79.2 & 71.04 \\
  HMNet & RGB+SM$_{DP}$ & 77.6 & 69.6\\
  HMNetOracle & RGB+SM$_{GT}$ & \textbf{42.9} & \textbf{37.9} \\
  \hline
\end{tabular}
\caption{Effect of each network module on the reconstruction error on UP-3D dataset. SM$_{DP}$ and SM$_{GT}$ denotes segmentation obtained from Densepose and groundtruth respectively.}
\label{table:ablation}
\end{table}

\begin{figure}[h!]
\includegraphics[width=0.9\linewidth]{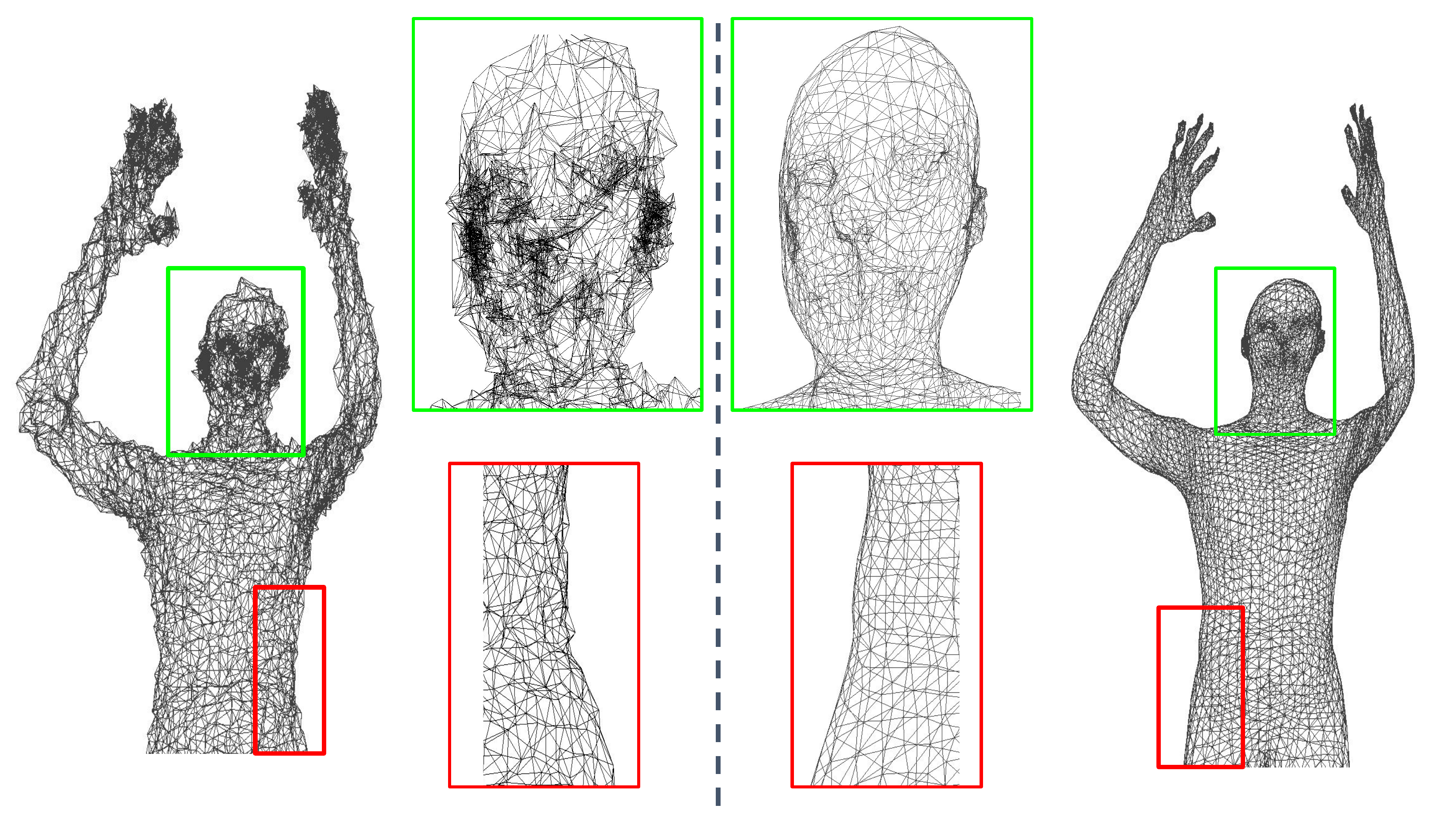}
\caption{Results showing the effect of our mesh regularization module while learning. The figure on the left shows the irregularities in the mesh reconstructed, without our regularization, while the one on the right shows the smoothness induced by our regularizer}
\label{fig:mesh_regularize}
\end{figure}

\noindent \textbf{Effect of Mesh Regularisation} Our mesh regularization module adds a smoothing effect while training, therefore ensuring that the entire local patch should move towards the ground truth for minimizing the error. Although intrinsic geometry based losses can also be used here, we hypothesize that they have a larger impact when more complex local surface deformations (e.g., facial expressions) are present. 
Figure~\ref{fig:mesh_regularize} show the impact of this regularization.

\begin{figure}[h!]
\begin{subfigure}[b]{0.15\linewidth}
\includegraphics[width=\linewidth]{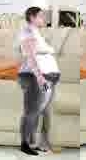}
\caption*{(a)}
\end{subfigure}
\begin{subfigure}[b]{0.15\linewidth}
\includegraphics[width=\linewidth]{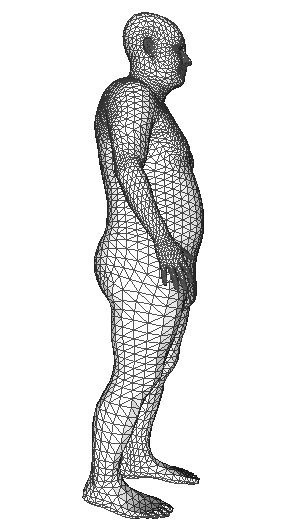}
\caption*{(b)}
\end{subfigure}
\begin{subfigure}[b]{0.15\linewidth}
\includegraphics[width=\linewidth]{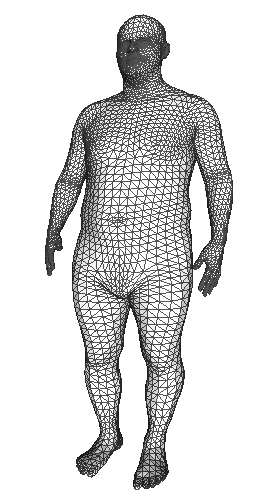}
\caption*{(c)}
\end{subfigure}
\begin{subfigure}[b]{0.15\linewidth}
\includegraphics[width=\linewidth]{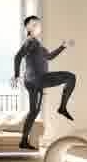}
\caption*{(a)}
\end{subfigure}
\begin{subfigure}[b]{0.15\linewidth}
\includegraphics[width=\linewidth]{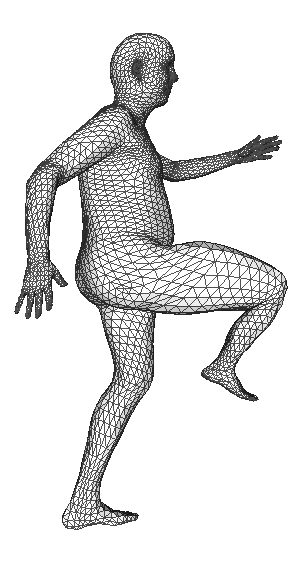}
\caption*{(b)}
\end{subfigure}
\begin{subfigure}[b]{0.15\linewidth}
\includegraphics[width=\linewidth]{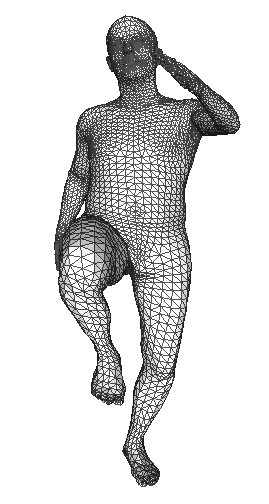}
\caption*{(c)}
\end{subfigure}
\caption{Sample Shape Variations recovered by our model given an input image (a), rendered from the recovered view (b) and another arbitary view (c)}
\label{fig:shapes}
\end{figure}

\noindent \textbf{Recovering Shape Variations } Most parametric models prediction work with a neutral template model~\cite{kanazawa2018end}, and would have to learn the gender from the image.  In our method, a direct mesh regression can learn the local shape variations (as long as training data has such variations) which extends to inherently learning gender invariant meshes. Two such samples are showing in Figure~\ref{fig:shapes}. \\

\begin{figure}[h!]
\begin{subfigure}[b]{0.15\linewidth}
  \includegraphics[width=\linewidth]{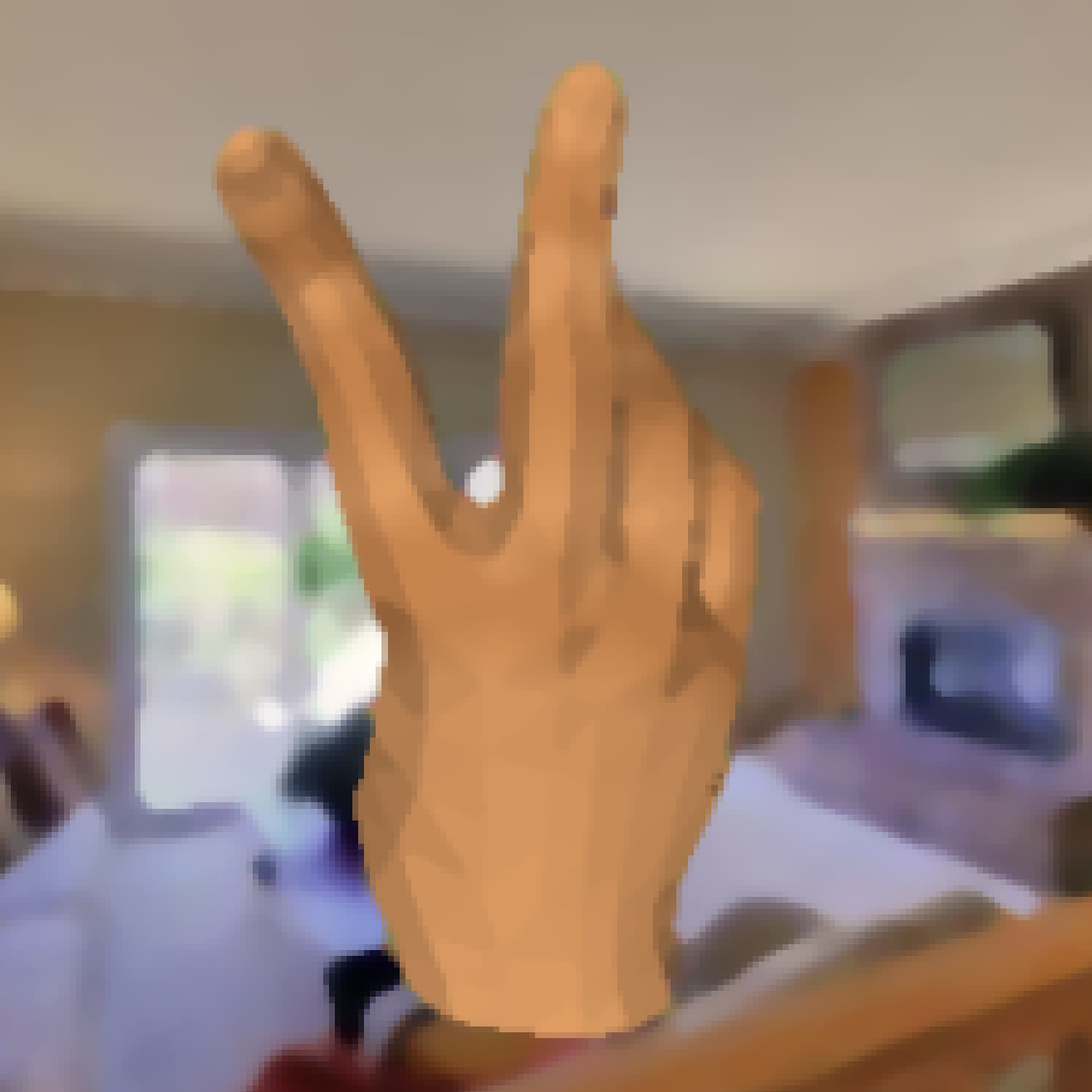}
\end{subfigure}
\begin{subfigure}[b]{0.15\linewidth}
\includegraphics[width=\linewidth]{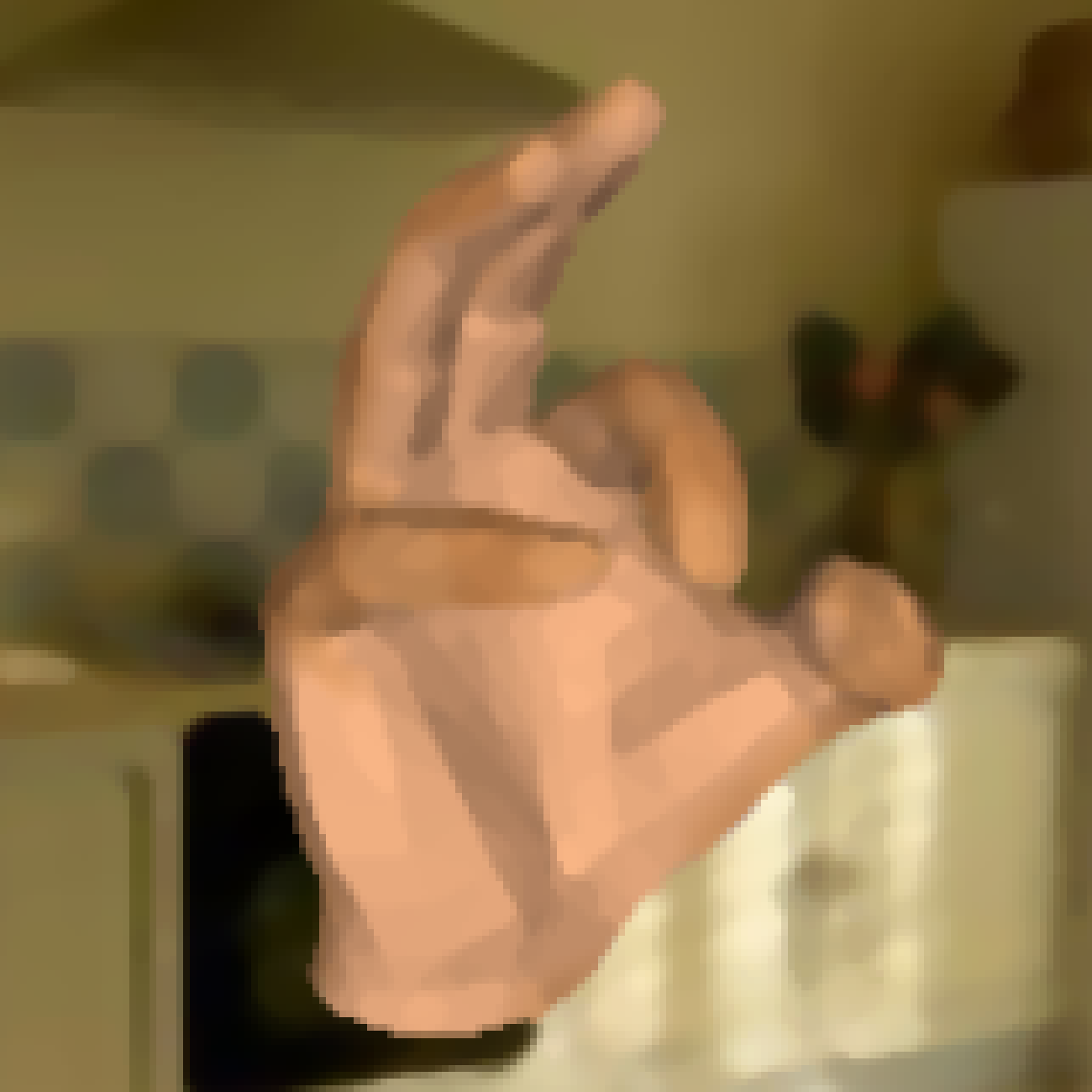}
\end{subfigure}
\begin{subfigure}[b]{0.15\linewidth}
\includegraphics[width=\linewidth]{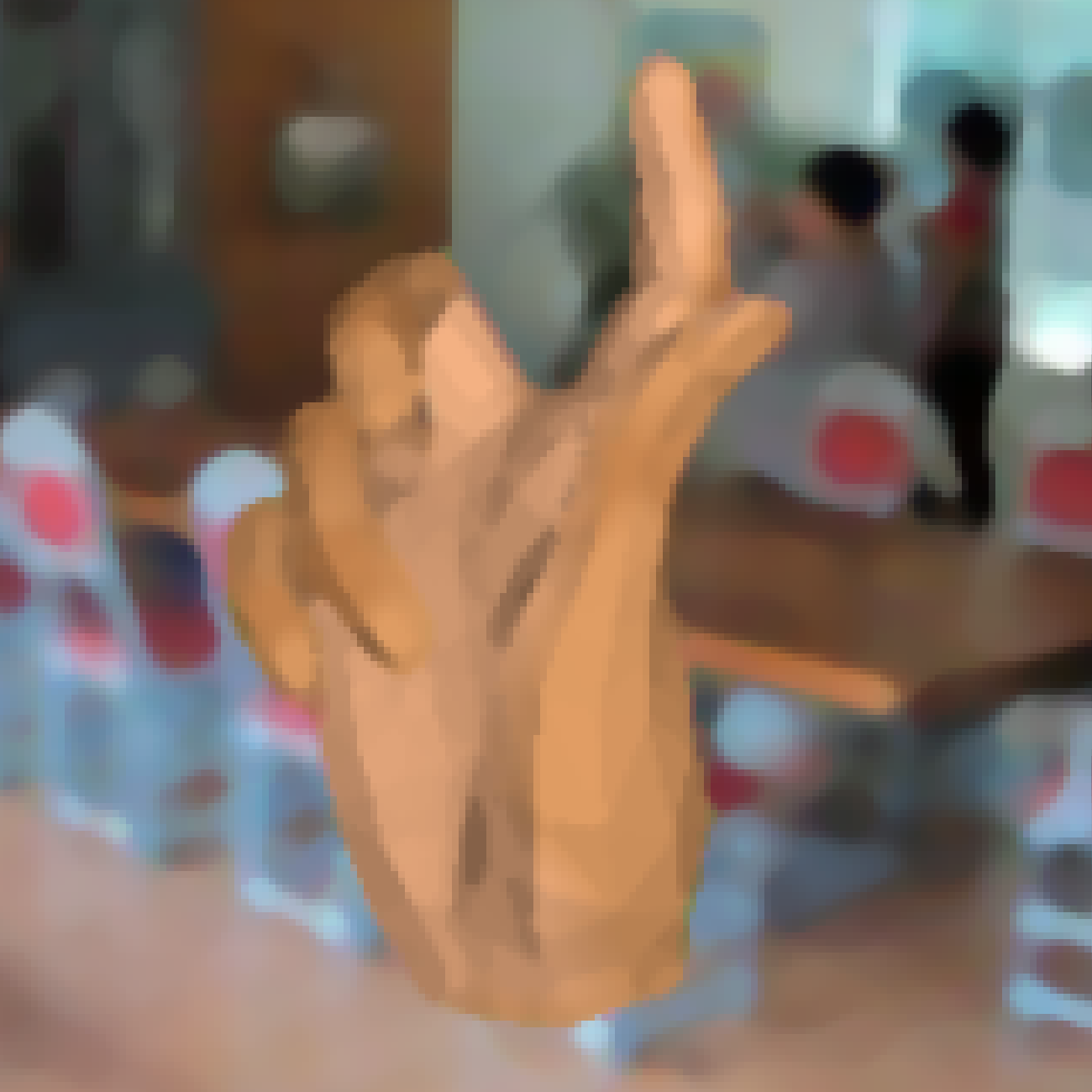}
\end{subfigure}
\begin{subfigure}[b]{0.15\linewidth}
\includegraphics[width=\linewidth]{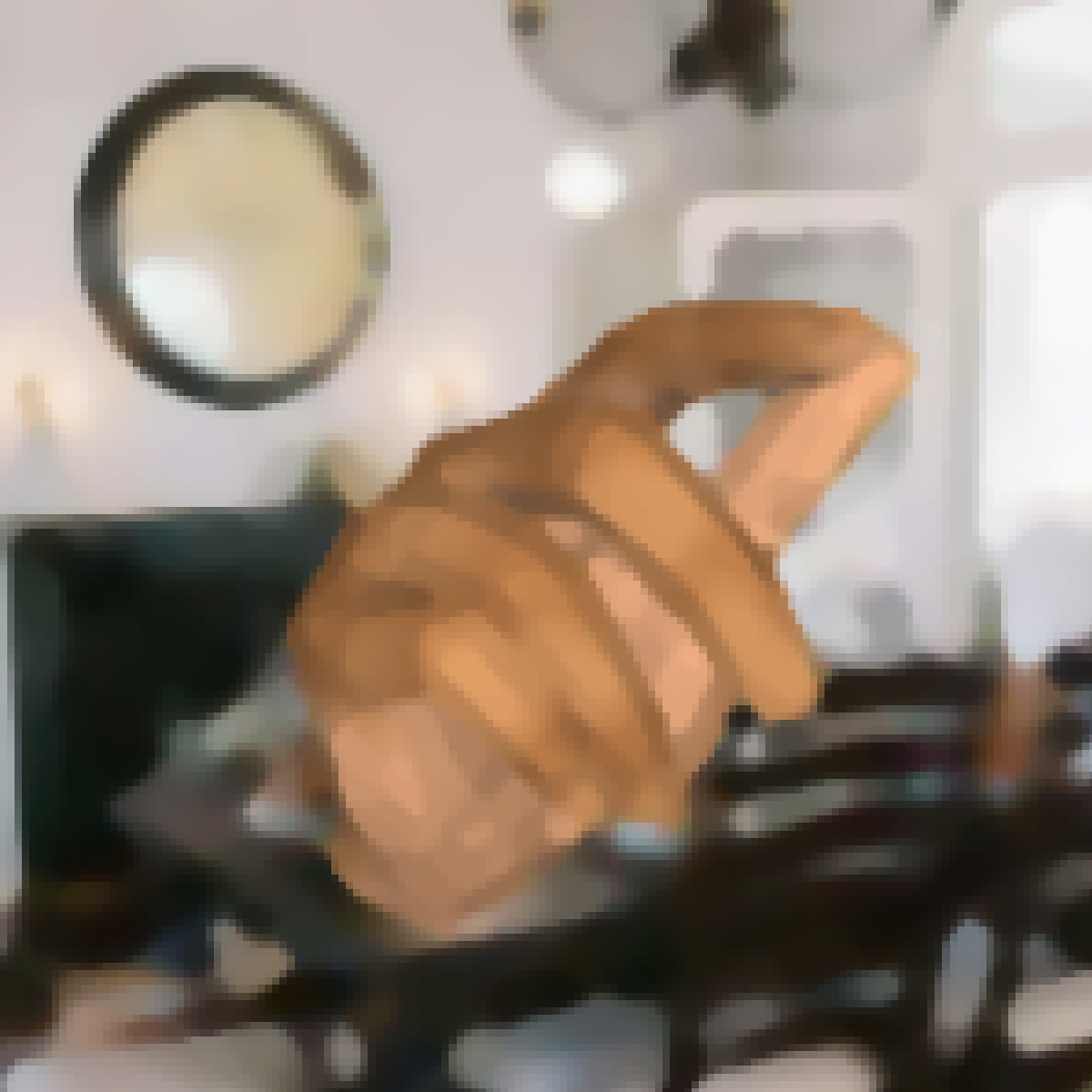}
\end{subfigure}
\begin{subfigure}[b]{0.15\linewidth}
\includegraphics[width=\linewidth]{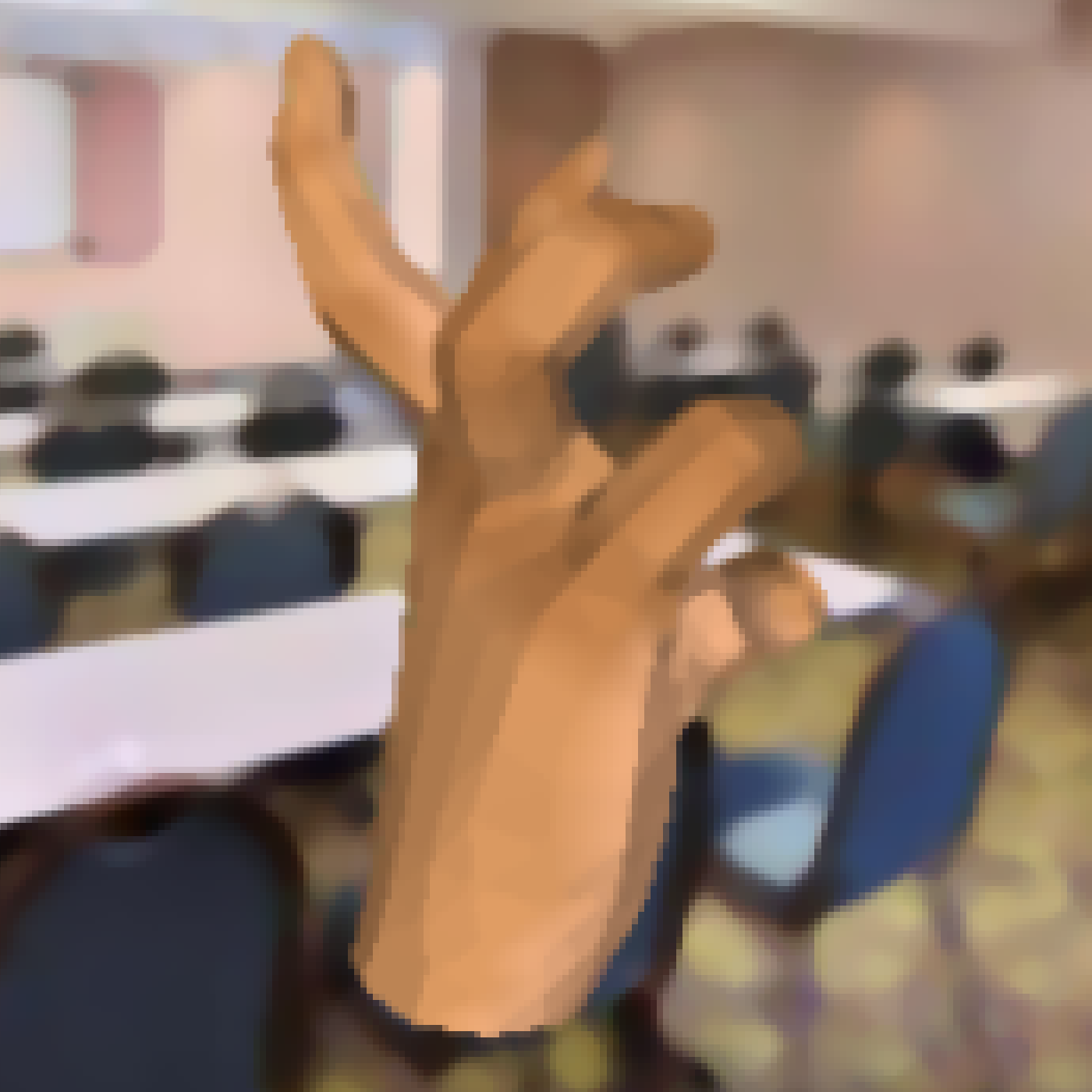}
\end{subfigure}
\begin{subfigure}[b]{0.15\linewidth}
\includegraphics[width=\linewidth]{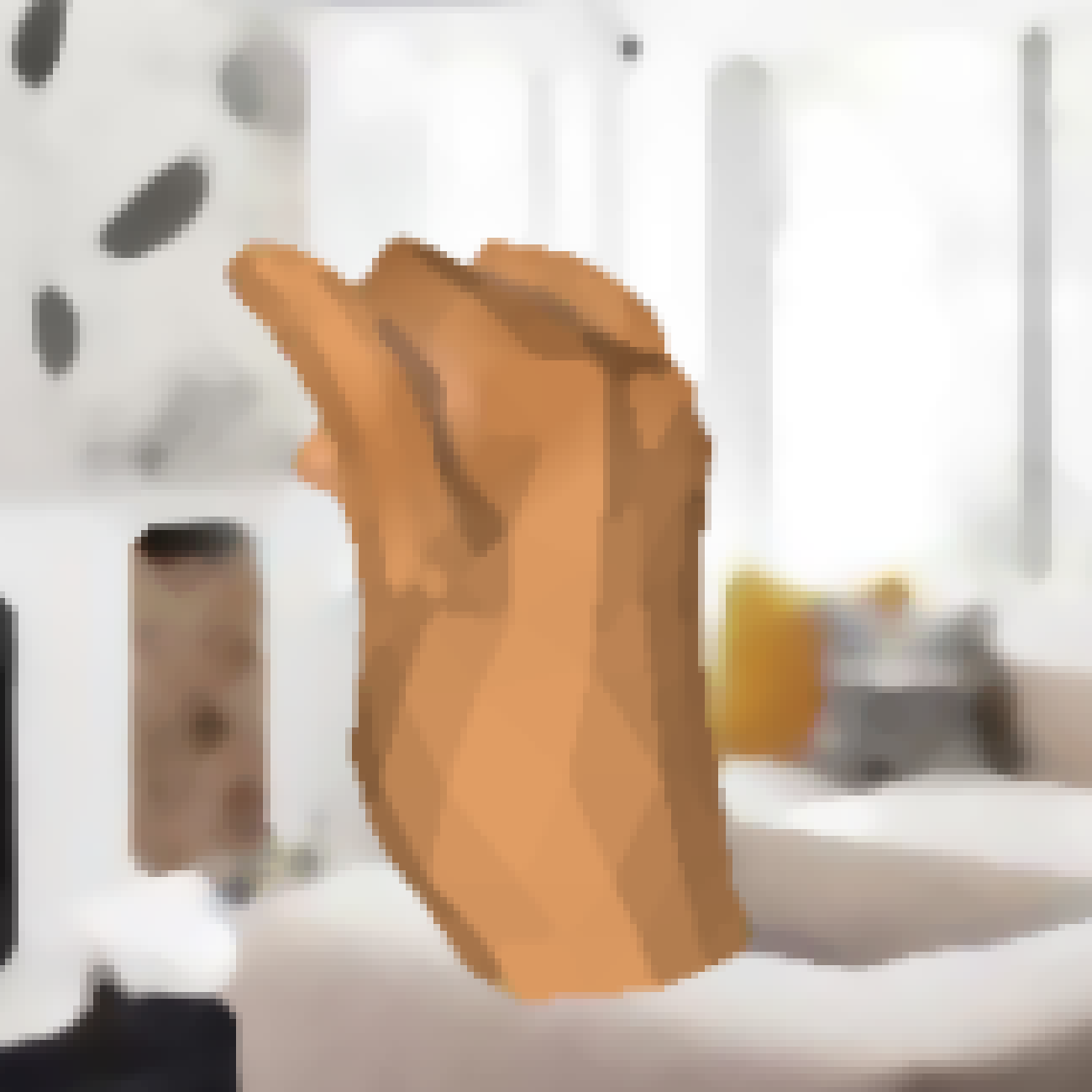}
\end{subfigure}

\begin{subfigure}[b]{0.15\linewidth}
\includegraphics[width=\linewidth]{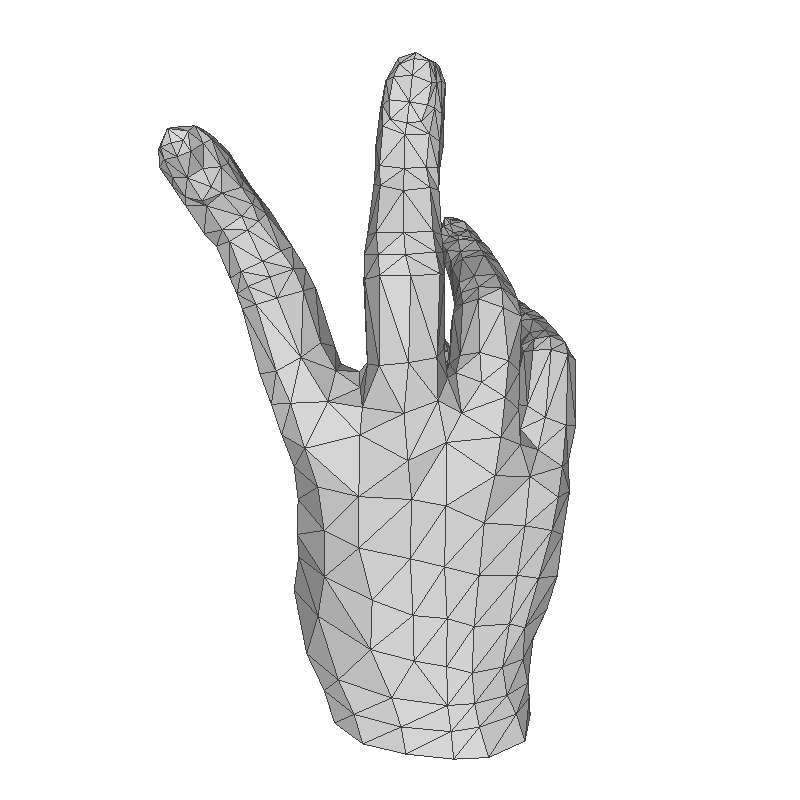}
\end{subfigure}
\begin{subfigure}[b]{0.15\linewidth}
\includegraphics[width=\linewidth]{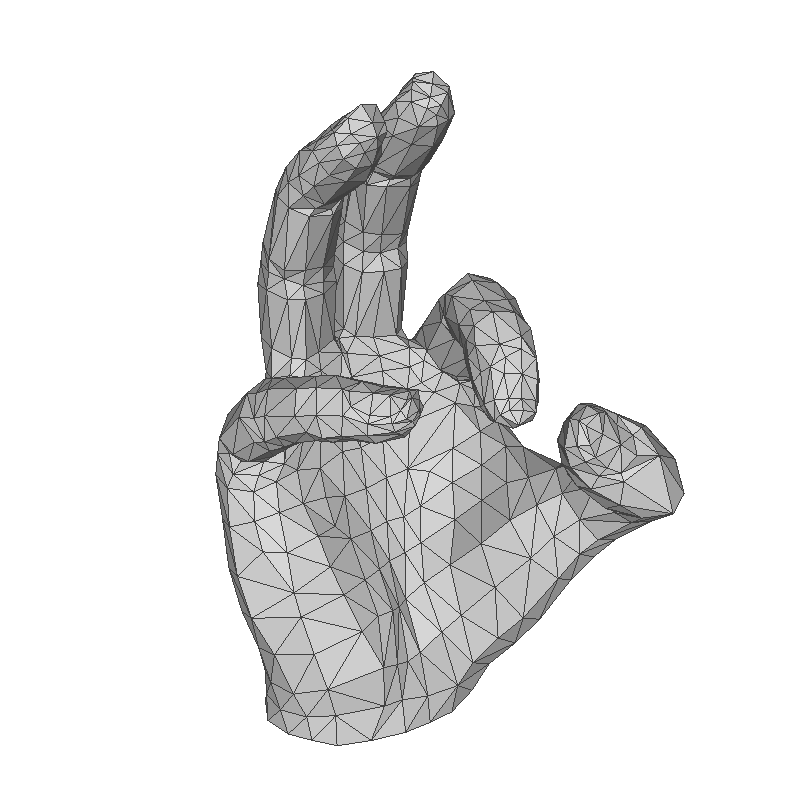}
\end{subfigure}
\begin{subfigure}[b]{0.15\linewidth}
\includegraphics[width=\linewidth]{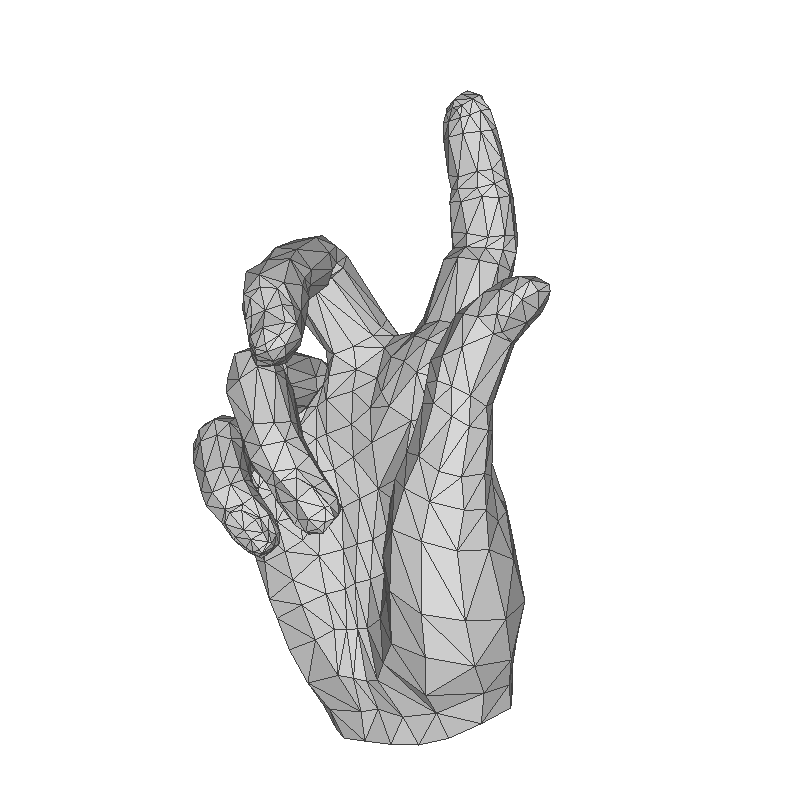}
\end{subfigure}
\begin{subfigure}[b]{0.15\linewidth}
\includegraphics[width=\linewidth]{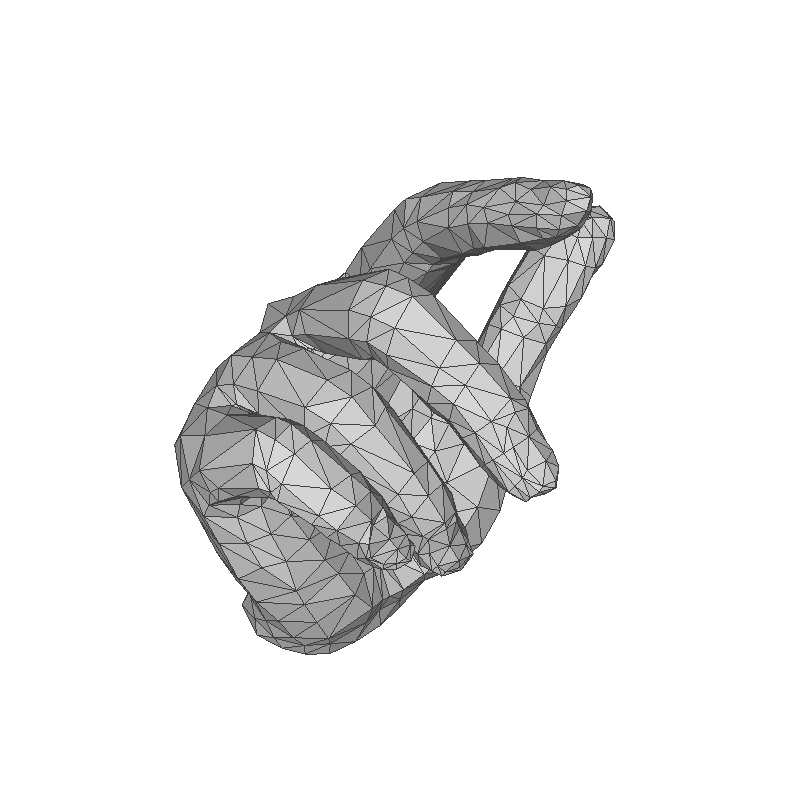}
\end{subfigure}
\begin{subfigure}[b]{0.15\linewidth}
\includegraphics[width=\linewidth]{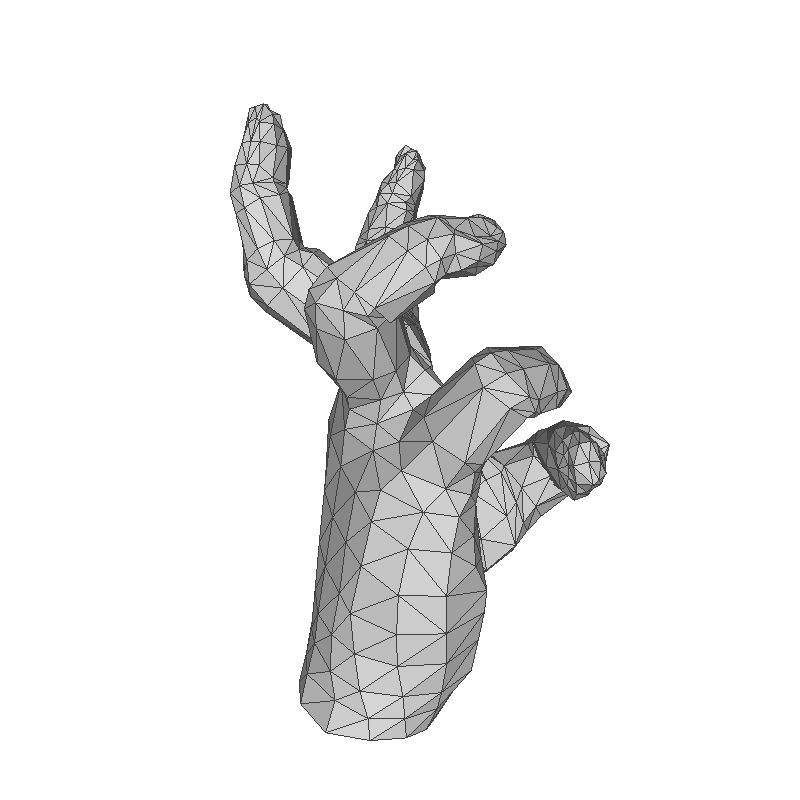}
\end{subfigure}
\begin{subfigure}[b]{0.15\linewidth}
\includegraphics[width=\linewidth]{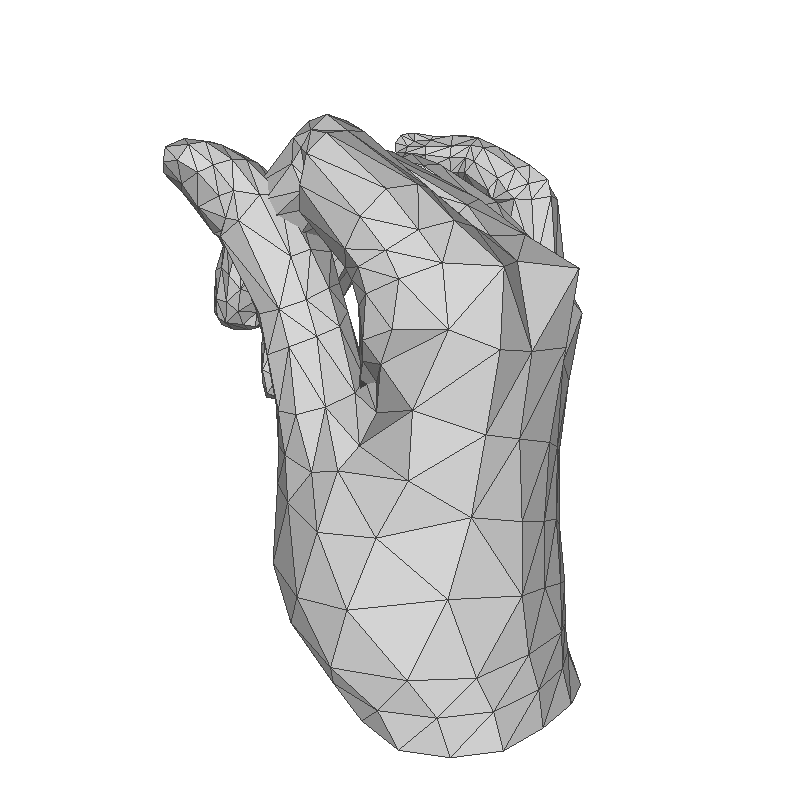}
\end{subfigure}

\begin{subfigure}[b]{0.15\linewidth}
\includegraphics[width=\linewidth]{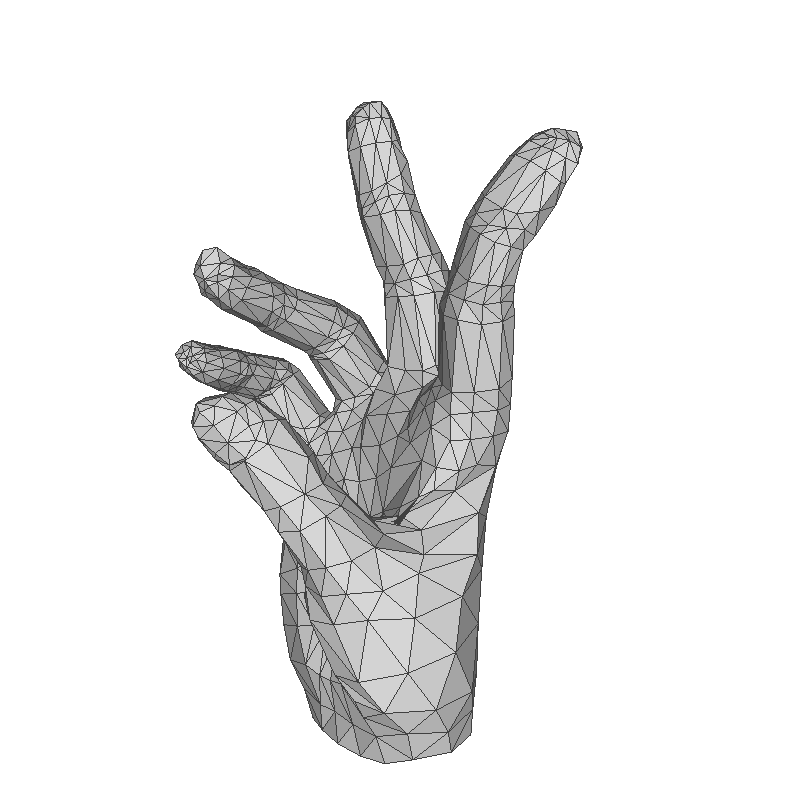}
\caption{}\label{fig:hand}
\end{subfigure}
\begin{subfigure}[b]{0.15\linewidth}
\includegraphics[width=\linewidth]{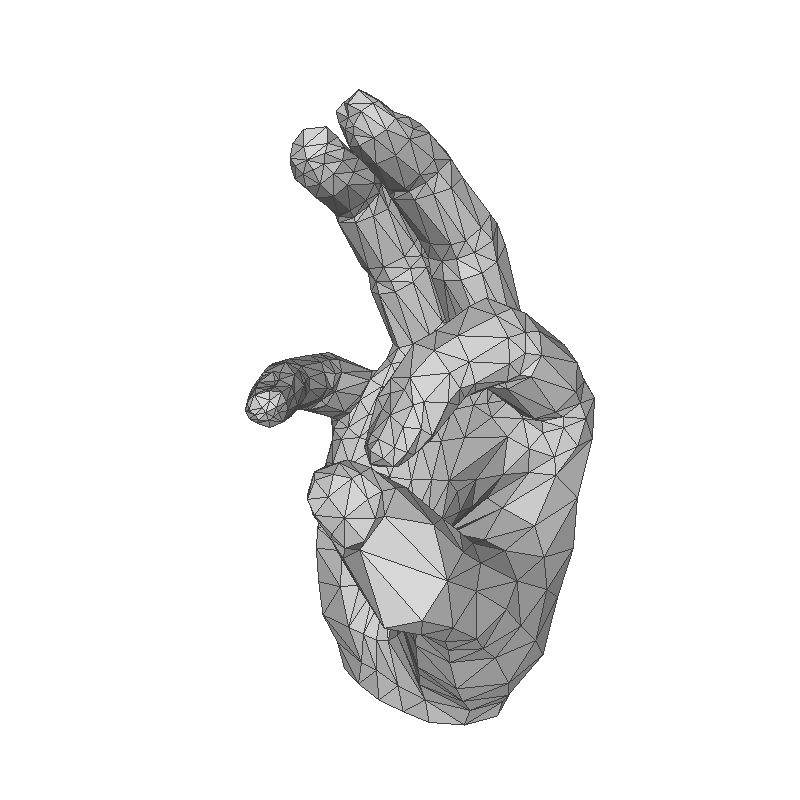}
\caption{}\label{fig:hand}
\end{subfigure}
\begin{subfigure}[b]{0.15\linewidth}
\includegraphics[width=\linewidth]{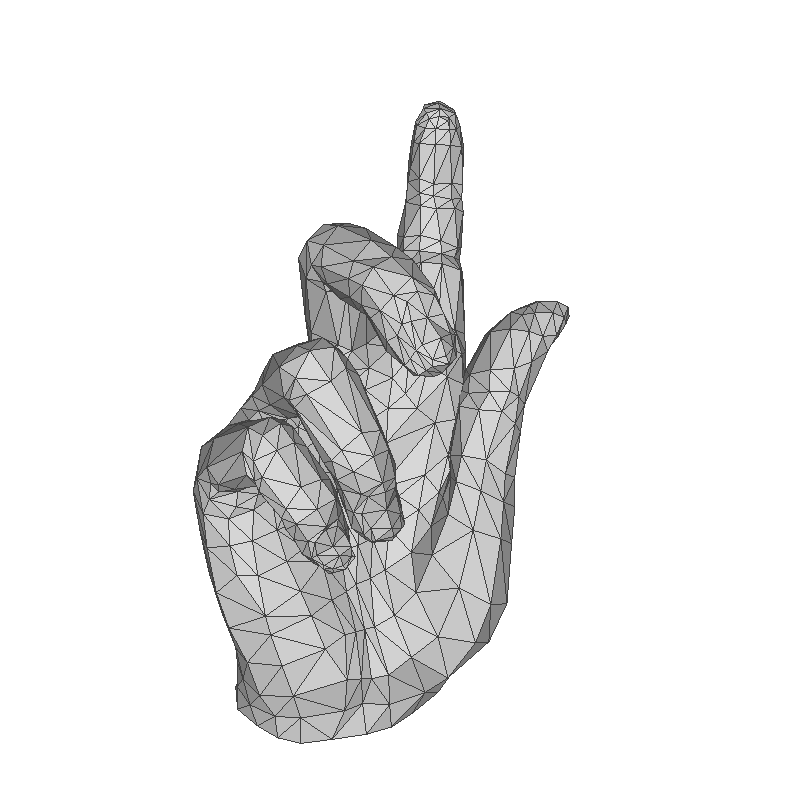}
\caption{}\label{fig:hand}
\end{subfigure}
\begin{subfigure}[b]{0.15\linewidth}
\includegraphics[width=\linewidth]{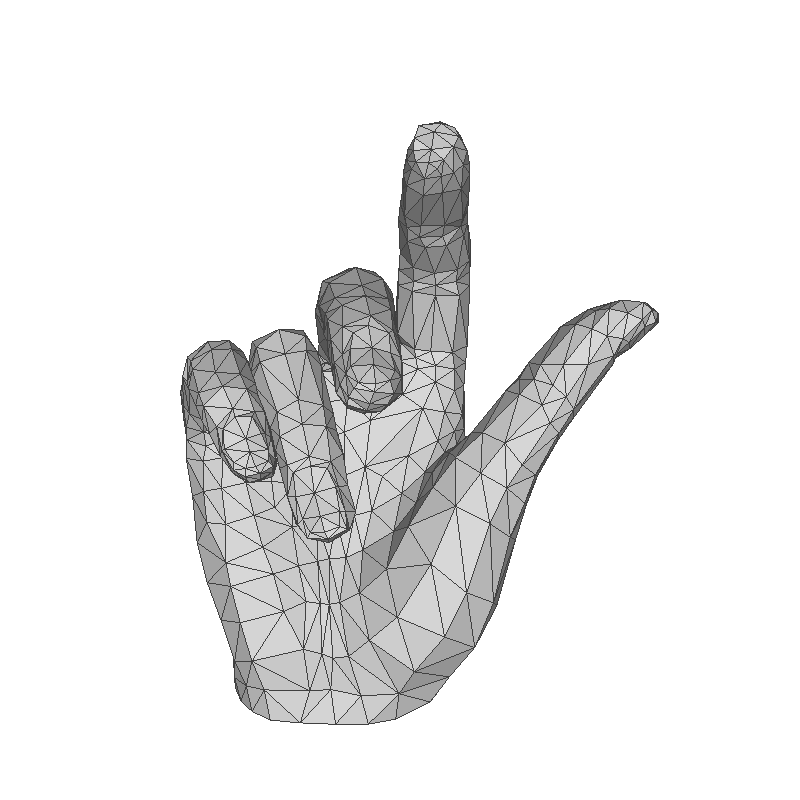}
\caption{}\label{fig:hand}
\end{subfigure}
\begin{subfigure}[b]{0.15\linewidth}
\includegraphics[width=\linewidth]{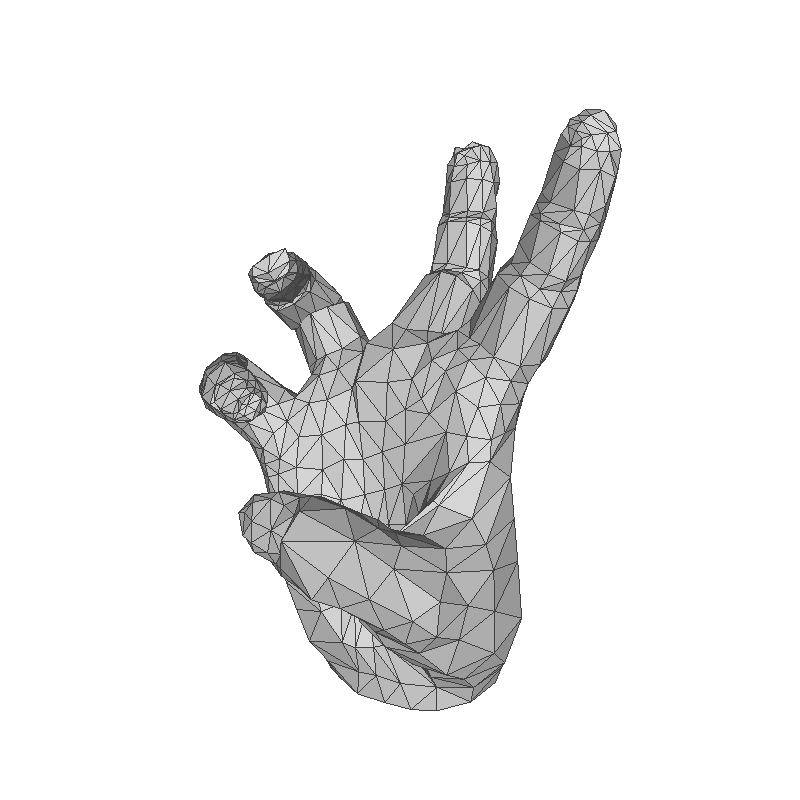}
\caption{}\label{fig:hand}
\end{subfigure}
\begin{subfigure}[b]{0.15\linewidth}
\includegraphics[width=\linewidth]{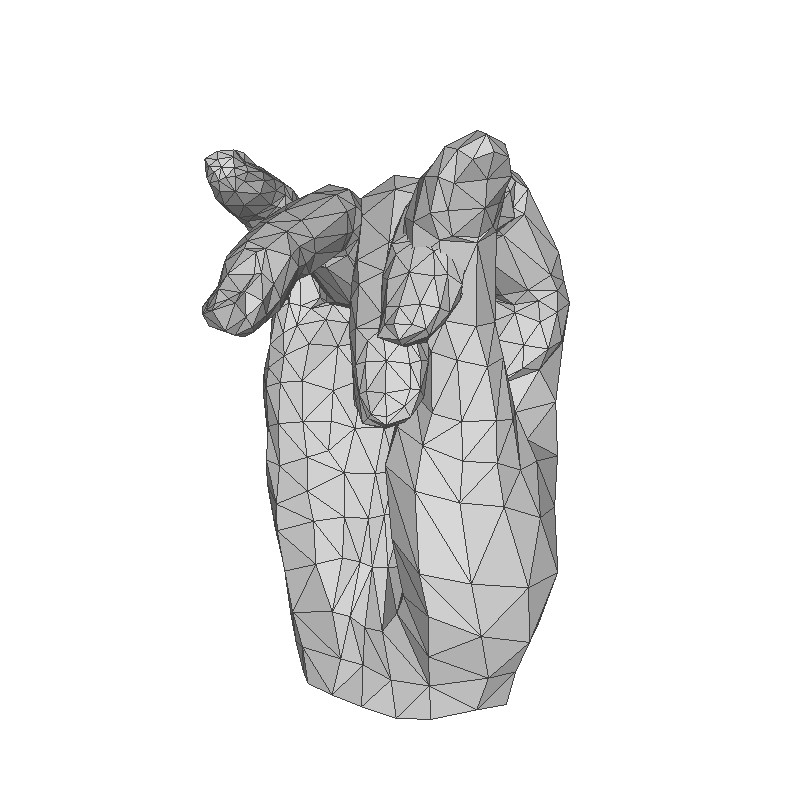}
\caption{}\label{fig:hand}
\end{subfigure}
\caption{Reconstruction Results on our Hand Mesh. Each column consists of the RGB image, its corresponding reconstruction from the same view, and from another arbitrary view. }
\label{fig:hand_res}
\end{figure}

\noindent \textbf{Generalizability to Hand Mesh Models.} We show the generalizability of our model to a similar task with a different strcuture. First, we populated a SURREAL like synthetic hand dataset using the MANO hand model ~\cite{romero2017embodied}, similar to ~\cite{ge20193d} with a total of 70,000 image-mesh pairs. We train our model on this dataset to predict hand surface and joints from an input RGB image using the same pipeline described in Figure~\ref{fig:arch}. The training setting remains the same as earlier, and we obtain impressive qualitative results as shown in Figure~\ref{fig:hand_res}. 
The average surface error across the test dataset is 1mm, which acts as a proof of concept that polygonal mesh reconstruction of non-rigid hands (although in a simplistic scenario), is feasible. \\

\noindent \textbf{Network Runtime.} Table~\ref{table:run_time} list out run-time of various methods.
Comparing this with HMNet with HMNetOracle, it is evident that a major part of HMNet's complexity arises from the multi-human pixel wise class prediction, which runs at around 30 FPS for an image of size 224x224.~\cite{charles2018real} is an accurate real time body part segmentation network which runs at 120 FPS, and can be incorporated into our system to produce accurate, real time reconstructions. \\
\begin{table}[h!]
\centering
\begin{tabular}{| c | c | c |}
    \hline
    Method & Output &  FPS \\
   \hline
    SMPLify~\cite{keepSMPL2016}& \multirow{3}{*}{P} & 0.01\\
    SMPLify, 91 kps~\cite{lassner2017unite} &  & 0.008 \\
    Decision Forests~\cite{lassner2017unite} &  & 7.69 \\
    \hline
    HMR~\cite{kanazawa2018end} & \multirow{3}{*}{P} & 25 \\
    Pavlakos~\cite{pavlakos2018learning} &  & 20 \\
    Direct Prediction~\cite{lassner2017unite} &  & 2.65 \\
    \hline
    Baseline & \multirow{3}{*}{S} & 175.4\\
    HMNet &   & \textbf{28.01}\\
    HMNetOracle &   & \textbf{173.17} \\
   \hline
    Fusion4D~\cite{Fusion4D:2016}& S & 31 \\
    \hline
\end{tabular}
\caption{Overview of the run time (in Frames Per Second, FPS) of various algorithms. Numbers have been picked up from the respective papers. All methods have used 1080Ti or equivalent GPU. }
\label{table:run_time}
\end{table}

\noindent \textbf{Limitations and Future Work.} Since we do not enforce any volume consistency, skewing/thinning artifacts might be introduced in our meshes. We would like to account for these in a non-handcrafted anthropomorphically valid way by either learning the SMPL parameters on top of it using an MLP similar to~\cite{kolotouros2019cmr} or by using a GAN to penalize fake/invalid human meshes. Further, we have made use of the mesh topology in two ways in this work - (a) implicitly, to make the learning easier and (b) for smoothing. Going ahead, we would like to make use of the mesh topology and geometry details is a more explicit manner, by using intrinsic mesh/surface properties. We believe that this is a largely unexplored space and applying such a regularization can result better exploitation of surface geometry for reconstruction.

\begin{figure}[h!]
\begin{subfigure}[b]{0.24\linewidth}
\begin{center}
\includegraphics[width=\linewidth]{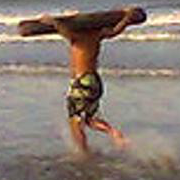}
\end{center}
\end{subfigure}
\begin{subfigure}[b]{0.24\linewidth}
\includegraphics[width=\linewidth]{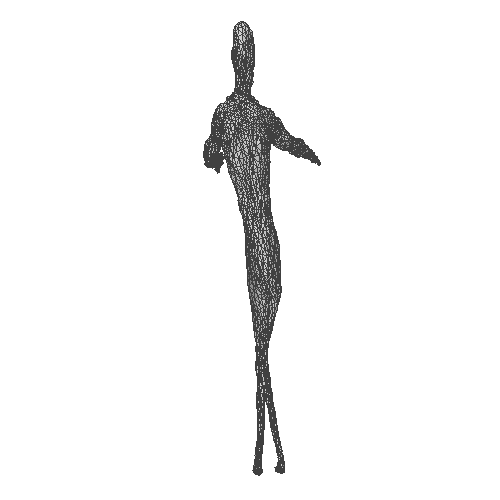}
\end{subfigure}
\begin{subfigure}[b]{0.24\linewidth}
\includegraphics[width=\linewidth]{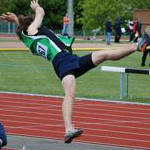}
\end{subfigure}
\begin{subfigure}[b]{0.24\linewidth}
\begin{center}
\includegraphics[width=\linewidth]{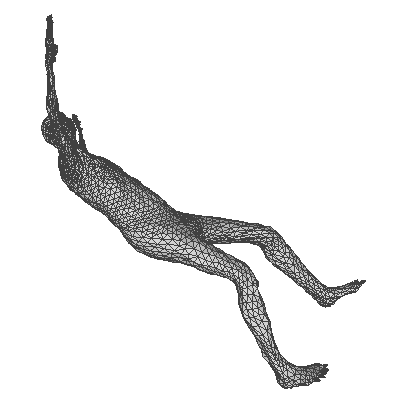}
\end{center}
\end{subfigure}
\caption{Failure Cases of Our Method. }
\label{fig:failure}
\end{figure}

\section{Conclusion}
We proposed a multi-branch multi-task HumanMeshNet network that simultaneously regress to the template mesh vertices as well as body joint locations from a single monocular image. The proposed method achieves comparable performance with significantly lower modelling and computational complexity on three publicly available datasets.  We also show the generalizability of the proposed architecture for similar task of predicting the mesh of the hand. Looking forward, we would like to exploit intrisinc mesh properties to recover a more accurate surface reconstruction.



{\small
\bibliographystyle{ieee}
\bibliography{egbib}
}

\end{document}